\documentclass[10pt,twocolumn,letterpaper]{article}
\ifdefined\pdfminorversion
\fi

\PassOptionsToPackage{table}{xcolor}
\usepackage[pagenumbers]{cvpr}
\usepackage{placeins}
\usepackage{multirow}
\usepackage{array}
\newcolumntype{B}{>{\Large\bfseries}c}
\usepackage{soul}
\usepackage{dsfont}

\definecolor{cvprblue}{rgb}{0.21,0.49,0.74}
\usepackage[pagebackref,breaklinks,colorlinks,allcolors=cvprblue]{hyperref}

\def\confName{CVPR}
\def\confYear{2027}

\title{Chain-of-Spatial-Thoughts: Architecture-Agnostic Spatial Grounding for Vision Language Models}

\author{Anonymous Authors}

\author{%
    Hunter Schofield\textsuperscript{1,2,*},\quad
    Mohammed Elmahgiubi\textsuperscript{2,*},\quad
    Mohammad Mahdavian\textsuperscript{2,*}\\
    Richard Shi\textsuperscript{2,3},\quad
    Jinjun Shan\textsuperscript{1},\quad
    Amir Rasouli\textsuperscript{2},\quad
    Dongfeng Bai\textsuperscript{2}\\[0.4em]
    \textsuperscript{1}York University\quad
    \textsuperscript{2}Huawei Technologies Canada\quad
    \textsuperscript{3}University of Toronto\\[0.3em]
    {\small \textsuperscript{*}Equal contribution.}\\
    {\ttfamily\small \{hunterls,jjshan\}@yorku.ca\quad rshi@cs.toronto.edu}\\
    {\ttfamily\small \{mohammed.elmahgiubi1,mohammad.mahdavian1,}\\
    {\ttfamily\small amir.rasouli,baidongfeng\}@huawei.com}%
}

\hypersetup{%
    pdftitle={\thetitle},
    pdfauthor={Hunter Schofield; Mohammed Elmahgiubi; Mohammad Mahdavian; Richard Shi; Jinjun Shan; Amir Rasouli; Dongfeng Bai}%
}

\begin{document}
\raggedbottom
\maketitle

\begin{abstract}
    Spatial understanding is fundamental to embodied intelligence, underpinning applications such as robotic manipulation, embodied navigation, and autonomous driving. Although recent vision-language models (VLMs) have achieved impressive performance on spatial reasoning benchmarks, state-of-the-art approaches typically rely on additional spatial encoders or architectural modifications during inference, increasing computational cost. 
    We introduce \textbf{Space Tokens}, a lightweight, architecture-agnostic framework that equips VLMs with explicit continuous spatial representations without requiring additional inference-time modules. By distilling scene-level 3D geometry and object-centric spatial attributes into continuous latent tokens, our method enables these modalities to be directly incorporated into a chain-of-thought reasoning process, thereby improving the VLM’s spatial reasoning capabilities. At the same time, the learned representations can be explicitly decoded to verify the meaningful geometric information, while the unified token interface remains extensible to additional modalities.
    Experiments on VSI-Bench improve InternVL3-8B by 13.8 percentage points, Qwen3-VL-8B by 4.3 percentage points and SenseNova-SI-1.3 by 1.4 percentage points, while achieving state-of-the-art performance on object size ($79.0\%$) and room size estimation ($76.7\%$). These results demonstrate that continuous spatial tokens provide an effective, interpretable, and computationally efficient mechanism for integrating geometric reasoning into large vision-language models. The project page is at: \url{https://mmahdavian.github.io/chain-of-spatial-thoughts/}
\end{abstract}

\section{Introduction}

\begin{figure}[!t] 
    \centering
    \includegraphics[width=\linewidth]{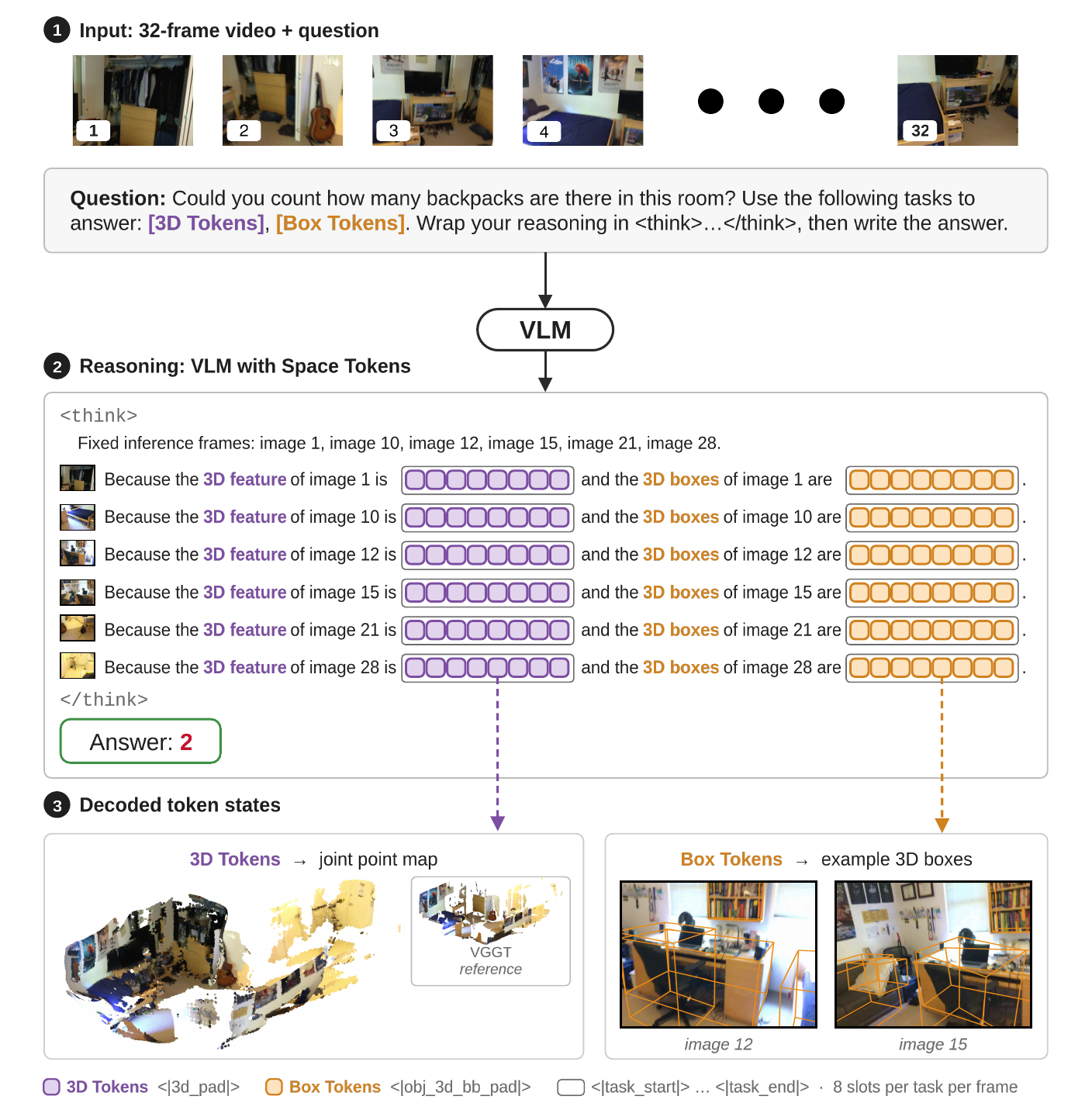} 
    \caption{Space Tokens Overview. Our approach embeds interpretable scene-level and object-centric spatial representations directly into the VLM for use during autoregressive reasoning. The learned tokens improve spatial reasoning and can be decoded into explicit spatial modalities to verify the model's learned geometric understanding} 
    \label{fig:teaserv2}
\end{figure} 

Recent advances in Multimodal Large Language Models (MLLMs), also referred to as VLMs, have enabled systems that jointly reason over visual observations and natural language \cite{Bai_qwenvl3, Li_blip2, Deng_bagel, gemini, gemma}. Beyond traditional image understanding, there is growing interest in equipping these models with the ability to understand the underlying 3D geometry, spatial relationships, and physical interactions depicted in images and videos \cite{Yang_vsi_bench, Fan_vlm3r}. Such spatial intelligence is fundamental to real-world applications including robotics, embodied agents, autonomous driving, and human-computer interaction, where successful decision making depends not only on recognizing objects, but also on understanding their geometric arrangement and interactions within the surrounding environment.

Progress in spatial reasoning has largely followed two complementary directions. First (1), \emph{foundational scaling} enhances spatial capabilities as a byproduct of expanding model parameter quantity and training datasets with internet-scale data \cite{Bai_qwenvl3, Deng_bagel, gemini} or by training on increasingly large spatially focused supervised datasets, as demonstrated by recent models such as SenseNova-SI \cite{Cai_sensenova}. 
Second (2), \emph{specialized spatial architectures}, introduces explicit spatial inductive biases through dedicated geometry modules, including 3D reconstruction encoders, geometry-aware connectors, object-centric representations, and scene-level relational reasoning \cite{Zhao_spacemind, Li_thinkwithgeometry, Chen_thinkwith3d, wang2025n3d, liu2025abstract, zhang2026ssr}. 
More recently, Latent Token-Based Knowledge Integration has emerged as a general methodology for incorporating auxiliary knowledge into VLMs through continuous latent token representations, allowing external representations to be distilled into the model while preserving the underlying architecture \cite{Qin_covt, Yang_mirage, Ray_mull}. Existing works have primarily leveraged this paradigm to internalize visual representations for improving perception and general 2D reasoning, leaving its potential for encoding explicit spatial and geometric knowledge largely unexplored.

Together, these approaches have advanced performance on benchmarks such as VSI-Bench \cite{Yang_vsi_bench}, VLM-3R \cite{Fan_vlm3r}, and EASI \cite{Cai_EASI}, yet key limitations remain. Scaling-based methods require large models or extensive spatial supervision, while explicit spatial approaches add inference-time modules that increase cost and reduce portability. Latent token methods avoid these drawbacks but have largely focused on visual or semantic knowledge, leaving explicit 3D geometry and object-centric representations underexplored. As a result, current VLMs still lag behind human spatial understanding, particularly on long-horizon tasks such as route planning and object appearance ordering \cite{Yang_vsi_bench}.

In this work, we build upon the paradigm of Latent Token-Based Knowledge Integration and extend it to spatial intelligence. Rather than scaling model capacity, scaling spatial supervision, or introducing large-size specialized inference-time geometry modules, we investigate whether explicit 3D geometric and object-centric spatial knowledge can instead be internalized within the existing VLM through lightweight training. As illustrated in Fig.~\ref{fig:teaserv2}, we introduce \textbf{Space Tokens}, a model-agnostic framework that distills scene-level geometry and object-centric spatial representations into continuous latent tokens that become part of the VLM's native chain-of-thought reasoning process. By extending latent token representations from visual understanding to explicit spatial knowledge, our method improves spatial reasoning while preserving the efficiency and portability of the underlying VLM. Furthermore, the learned representations remain directly decodable into interpretable 3D modalities, enabling explicit verification that the model has acquired meaningful geometric understanding without requiring architectural modifications or additional inference-time modules.

Our main contributions are summarized as follows:
\begin{itemize}
    \item We introduce Space Tokens, an architecture-agnostic and visually verifiable method that equips VLMs with continuous spatial representations through a unified token interface during chain-of-thought reasoning, requiring no additional inference-time modules or architecture modifications.
    \item We improve scene-level 3D understanding by distilling geometric knowledge from large 3D reconstruction models directly into the VLM’s latent space, removing the need for these models at inference time.
    \item We also enhance object-centric spatial reasoning by encoding 3D bounding-box information, including object position, orientation, and size, into dedicated latent tokens.
    \item We demonstrate the effectiveness of Space Tokens on several spatial, visual and general VQA benchmarks. Our method improves InternVL3-8B by 13.8 percentage points, Qwen3-VL-8B by 4.3 percentage points and SenseNova-SI-1.3 by 1.4 percentage points on VSI-Bench while achieving state-of-the-art performance on object size and room size estimation. 
\end{itemize}

\section{Related Work}

\subsection{Spatial Foundational Scaling.}

Recent improvements in spatial reasoning have been driven in part by scaling the capacity of general-purpose vision-language models and the amount of spatially supervised training data. Large multimodal foundation models such as Qwen3-VL, Gemini, and BAGEL demonstrate strong zero-shot and few-shot spatial reasoning capabilities that emerge from large-scale multimodal pretraining \cite{Bai_qwenvl3, Deng_bagel, gemini}. Complementing model scaling, SenseNova-SI \cite{Cai_sensenova} shows that scaling high-quality spatial instruction data substantially improves spatial reasoning across a wide range of tasks. Although these approaches establish increasingly capable general-purpose models, they typically require substantially larger models or datasets to achieve continued improvements.

\subsection{Explicit Spatial Representations.}

A second line of work improves spatial reasoning by explicitly introducing geometric representations into the VLM. Representative approaches include augmenting the VLM with dedicated geometry modules such as the camera-guided fusion architecture of SpaceMind \cite{Zhao_spacemind}, geometry encoders in GeoThinker and VG-LLM \cite{Li_thinkwithgeometry, Zheng_vgllm}, and reconstruction-guided reasoning frameworks such as VLM-3R \cite{Fan_vlm3r}. Beyond scene-level geometry, several methods incorporate object-centric spatial representations. LocateAnything3D~\cite{man2025locateanything3d} formulates 3D object detection as next-token prediction through a Chain-of-Sight procedure, while N3D-LM~\cite{wang2025n3d} jointly learns 3D grounding and spatial reasoning using 3D-aware QA supervision. SandboxVLM~\cite{liu2025abstract} represents scenes through abstract 3D bounding boxes, and SSR~\cite{zhang2026ssr} combines object-centric 7D representations with lightweight scene graphs. Chen et al.~\cite{chen2026thinking} further unify 6D object parsing, tracking, and geometric prediction before reinforcement learning refinement. Collectively, these methods demonstrate that explicit geometric representations significantly improve spatial reasoning; however, they generally rely on specialized architectural components or external geometry modules that remain active during inference, increasing computational cost and reducing portability across VLM backbones.

\subsection{Latent Token-Based Knowledge Integration.}

Rather than introducing additional architectural components, recent works have explored continuous latent tokens as a general mechanism for incorporating auxiliary knowledge into VLMs. Mirage \cite{Yang_mirage} introduces latent visual tokens that represent the model's internal mental imagery, enabling reasoning over visual concepts without explicitly generating intermediate images. 


Chain-of-Visual-Thought (CoVT) \cite{Qin_covt} instead focuses on enriching visual perception by distilling complementary information from multiple pretrained expert models, including depth estimation, semantic segmentation, and edge detection, into continuous latent representations that improve fine-grained image understanding. Collectively, these works establish latent tokens as an effective paradigm for internalizing auxiliary knowledge while preserving the underlying VLM architecture. However, existing approaches primarily employ latent tokens to encode visual representations that support perception and general 2D reasoning, leaving the incorporation of explicit 3D geometric and object-centric spatial knowledge largely unexplored.


Our work extends latent token-based knowledge integration from visual perception to explicit spatial intelligence. We embed complementary scene-level and object-centric 3D information into continuous latent tokens, enabling VLMs to internalize spatial priors within their native autoregressive reasoning process without architectural changes or additional inference-time modules. More broadly, latent tokens provide a unified interface for integrating structured spatial knowledge into foundation VLMs.
\section{Methodology}

\begin{figure*}[!ht] 
    \centering
    \includegraphics[width=\linewidth]{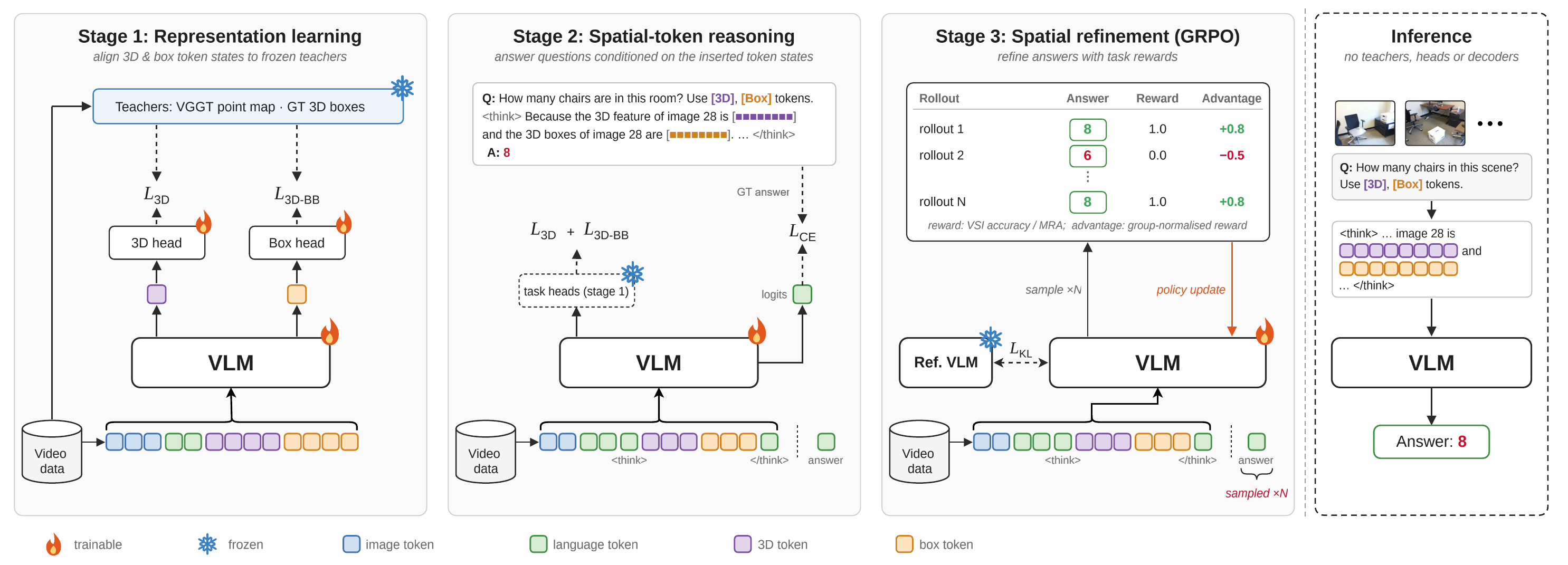} 
    \caption{Overview of the proposed three-stage Space Tokens training pipeline. \textbf{Stage 1} learns continuous spatial token representations by distilling scene-level 3D geometry from VGGT and object-centric spatial attributes into reserved output vocabulary tokens. \textbf{Stage 2} supervises the VLM to reason with the learned spatial tokens through standard next-token prediction SFT. \textbf{Stage 3} further refines spatial reasoning using reinforcement learning. After stage 1, all spatial reasoning is performed using the learned latent tokens without requiring additional model parameters.}
    \label{fig:pipeline}
\end{figure*} 

\subsection{Continuous Spatial Tokens}
To preserve the underlying autoregressive generation paradigm of the VLM and ensure that our approach remains lightweight and model agnostic, we introduce spatial tokens by reserving a subset of the vocabulary, without requiring any architectural modifications. Given a set of images or video frames $\mathcal{I} = \{ I_{1}, I_2, ..., I_n  \}$, a prompt $\mathcal{P}$, and model $M_{\theta}$, the output sequence $\mathcal{Y}$ is generated according to

\begin{equation}
    M_{\theta}(\mathcal{Y}|\mathcal{I}, \mathcal{P}) = \prod_{t=1}^{|\mathcal{Y}|}M_{\theta}(y_t|\mathcal{I}, \mathcal{P}, y_{<t})
\end{equation}

During autoregressive generation, spatial tokens are produced identically to any other tokens. Rather than using their discrete token identities, we extract the final layer hidden states corresponding to the generated positions of the spatial tokens. These serve as continuous spatial representations for downstream supervision.

Based on these representations, we define continuous spatial tokens that encode specific spatial concepts such as scene-level 3D reconstruction and instance-level 3D bounding boxes. Unlike conventional visual tokens that primarily capture semantic information, these tokens are optimized to represent fine-grained spatial structure enabling the VLM to reason about the geometry of its environment. To learn these representations, we supervise the spatial tokens using task-specific objectives. When a teacher model is available, spatial tokens are aligned with the teacher through feature-level and representation-level objectives, similar to CoVT \cite{Qin_covt}. For spatial concepts without an appropriate teacher model, spatial tokens are learned solely through the supervised task objective.

\subsubsection{3D Representation Tokens}
We align the 3D representation tokens with the latent scene representation learned by VGGT \cite{Wang_VGGT}. The objective is twofold. First, we encourage the VLM latent tokens to encode the same geometric information as the teacher through latent space alignment. Second, we ensure the representations remain geometrically meaningful by supervising their decoded 3D outputs. Accordingly, the 3D representation tokens are jointly optimized with similarity and multi-task reconstruction losses. The latent space objective encourages the continuous spatial tokens to capture the teachers internal geometric representation, while the reconstruction objectives ensure that these representations remain predictive of explicit 3D scene structure. 

For latent space alignment, we project the hidden states corresponding to the 3D reconstruction tokens into the VGGT latent space, $\hat{z} = \textrm{proj}(h_{3D})$, where $h_{3D}$ denotes the final-layer hidden states of the VLM backbone associated with the output 3D tokens. We then minimize the cosine distance between the projected representation, $\hat{z}$, and the corresponding VGGT aggregation features, $z$:

\begin{equation}
    \label{eq:sim}
    \mathcal{L}_{sim} = 1 - \frac{\hat{z}^{\top} z}{\left\Vert \hat{z} \right\Vert \left\Vert z \right\Vert}
\end{equation}

To further encourage the latent tokens to encode complete scene geometry, we additionally supervise the decoded outputs using the reconstruction objectives proposed by VGGT. Specifically, we reconstruct camera parameters, dense depth maps, and 3D point maps. The camera reconstruction objective, $\mathcal{L}_{cam} = \sum_{i=1}^N |\hat{\textbf{g}}_i - \textbf{g}_i|$, is an $\ell_1$-objective between the predicted camera pose encoding vector, $\hat{\textbf{g}}_i$, and the teacher estimation, $\textbf{g}_i$. Following the VGGT implementation, the depth reconstruction objective combines a regression term with a gradient consistency term, both weighted by an aleatoric uncertainty map, $\mathcal{L}_{dpt} = \sum_{i=1}^N \left\Vert c_{i}^D \odot \left( \hat{D}_i - D_i \right) \right\Vert + \left\Vert c_{i}^D \odot \nabla \left( \hat{D}_i - D_i \right) \right\Vert - \alpha \log(c_{i}^D)$, where $D_i$ and $\hat{D}_i$ denote the teacher and predicted depth maps, respectively, and $c_i^D$ represents the predicted uncertainty. The predicted depth maps are unprojected into camera coordinates using the estimated camera intrinsics to obtain $\hat{P}$. The resulting point maps are supervised analogously to the depth objective, $\mathcal{L}_{pt} = \sum_{i=1}^N \left\Vert c_{i}^D \odot \left( \hat{P}_i - P_i \right) \right\Vert + \left\Vert c_{i}^D \odot \nabla \left( \hat{P}_i - P_i \right) \right\Vert$ except that the aleatoric uncertainty term is omitted, as its effect is already accounted for in the depth supervision. Additional methodological details of the 3D reconstruction objective formulation are provided in the appendix. 

The complete training objective for the 3D reconstruction tokens is:

\begin{equation}
    \mathcal{L}_{3D} = \lambda_{sim} \mathcal{L}_{sim} + \lambda_{cam} \mathcal{L}_{cam} + \lambda_{dpt} \mathcal{L}_{dpt} + \lambda_{pt} \mathcal{L}_{pt}
    \label{eq:3d_rep_loss}
\end{equation}

\noindent where the $\lambda$ coefficients balance the contribution of each objective. The values used in our experiments are provided in the appendix. 

\subsubsection{3D Bounding Box Tokens}
To enhance the model’s object-centric understanding of the environment, we introduce a set of dedicated tokens for predicting object-level 3D bounding boxes. We adopt this representation because a 3D bounding box compactly encodes an object’s position, orientation, and spatial extent. These properties provide useful geometric priors for answering questions involving object dimensions, relative poses, spatial relationships, and scene layout.

To predict 3D bounding boxes, we combine the hidden representations of the dedicated bounding-box tokens with visual features extracted by the VLM’s vision backbone. We additionally incorporate features derived from the corresponding normalized 2D bounding boxes to provide object-localization guidance during training. As discussed previously, the auxiliary modalities do not necessarily need to be decoded at inference time. Consequently, the 2D bounding boxes are used only as training-time supervision and are not required during inference. 

We formulate 3D bounding-box prediction as fixed-cardinality set regression.
Each box is represented by a 12D vector containing its camera-space center,
dimensions, and 6D rotation representation. Predictions are matched to
ground-truth boxes using Hungarian assignment with an $\ell_1$ matching cost.
The bounding-box objective is defined as
\begin{equation}
\mathcal{L}_{\mathrm{3D\text{-}BB}}
=
\lambda_{\mathrm{obj}}\mathcal{L}_{\mathrm{obj}}
+
\lambda_{\mathrm{bg}}\mathcal{L}_{\mathrm{bg}},
\qquad
\label{eq:3dbb}
\end{equation}
\noindent where $\mathcal{L}_{\mathrm{obj}}$ is the Smooth $\ell_1$ loss
($\beta=0.1$) over Hungarian-matched boxes, and
$\mathcal{L}_{\mathrm{bg}}$ regresses unmatched slots toward a zero background target. The loss is summed over prediction slots and averaged over annotated views. The complete decoder architecture and objective are provided in the appendix.

\subsection{Three-Stage Pipeline}

The full training pipeline of the Space Tokens method is illustrated in Fig. \ref{fig:pipeline}. To adequately learn both the spatial representations and how to effectively use these representations during the reasoning process, we design a three-stage training pipeline. The first stage is the representation learning stage, in this stage we manually inject the spatial tokens in the prompt, directly embedded within a chain-of-thought style reasoning sentence. We compute the alignment losses using Eq. \ref{eq:3d_rep_loss} and Eq. \ref{eq:3dbb} by projecting the hidden states associated with the spatial tokens into their corresponding latent representations. In this stage, we only backpropagate these reconstruction losses to ensure that the projection layers learn a suitable mapping from the VLM hidden state to the spatial feature latent space, without having to compete with downstream tasks.

The second stage is the spatial-token reasoning stage. During this stage, we freeze the projection layers that map the VLM hidden states to the spatial-feature latent space, preventing the learned spatial representations from drifting during training. As in the first stage, the spatial tokens are retained in the prompt. We additionally optimize a cross-entropy objective under teacher forcing to encourage the VLM to explicitly incorporate the spatial tokens into its reasoning process. Importantly, our empirical results indicate that reasoning over a subset of the input frames is more effective than reasoning over the full set. Additional details regarding the prompt structure and frame-selection strategy are provided in the appendix. We also retain the spatial-alignment losses during this stage as regularization objectives. This helps preserve the spatial representations learned in the first stage while the model is optimized to produce the correct final answer.

The final stage is a spatial refinement stage. We perform group relative policy optimization (GRPO) \cite{Shao_GRPO} to further improve the ability of the model to reason with spatial representations. For a image set and prompt pair, $\left( \mathcal{I} , \mathcal{P} \right)$, we generate a group of $G$ output sequences, $\left\{ \mathcal{Y}_{1}, \ldots, \mathcal{Y}_{G} \right\}$ from the current model $M_{\theta_{\textrm{old}}}$. Using a scoring function to determine a reward for each sequence, we can take a policy gradient step to update $M_{\theta_{\textrm{old}}}$ to $M_{\theta}$ by optimizing the following cost


\begin{multline}
    \small
    \mathcal{J}(\theta) = \frac{1}{G}\sum_{i=1}^G \frac{1}{\left| \mathcal{Y}_i \right|} \sum_{t=1}^{\left| \mathcal{Y}_i \right|} \Big\{ \textrm{min} \Big[ r_{i, t} \hat{A}_{i, t},  \\
    \textrm{clip} \Big(r_{i,t}, 1 - \epsilon, 1 + \epsilon \Big) 
    \quad \hat{A}_{i, t} \Big] - \beta \mathbb{D}_{KL}\Big[ M_{\theta} || M_{\textrm{ref}} \Big] \Big\}
\end{multline}

\noindent where $r_{i,t} = \frac{M_{\theta}(y_{i,t}|y_{i, <t})}{M_{\theta_{\textrm{old}}}(y_{i,t}|y_{i, <t})}$ is the likelihood ratio for sampling the token from sequence $\mathcal{Y}_i$ at step t, and $\hat{A}_{i, t} = \frac{R_{i, t} - \bar{R}_t}{\sigma_{R}}$ is the group-normalized advantage for some reward, $R$. $\mathbb{D}_{KL}$ represents the Kullback-Leibler divergence which penalizes the model from diverging from the stage 2 reference model, and $\epsilon$ and $\beta$ are hyperparameters that control the magnitude of policy deviation and influence of the KL divergence penalty, respectively. Details on the reward design are included in the appendix.

\begin{figure}[t] 
    \centering
    \includegraphics[width=\linewidth]{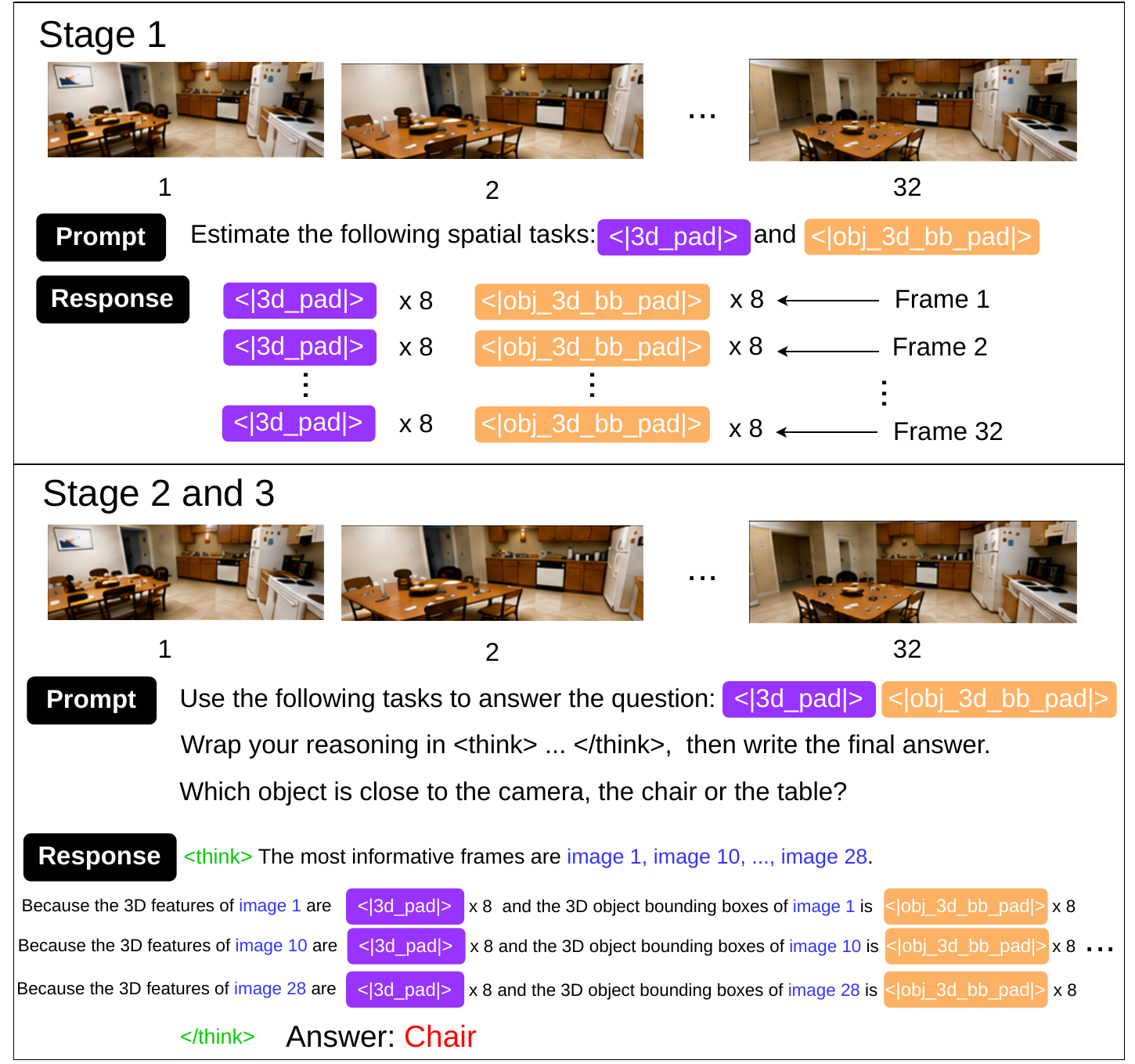} 
    \caption{Prompt formats used across training stages. Top (Stage 1): We perform representation learning by training the dedicated slots to encode spatial information. Bottom (Stages 2 and 3): We leverage the learned representations to improve the model’s spatial reasoning, first through SFT and then through GRPO training. }

    \label{fig:prompts}
\end{figure} 

\section{Experiments}
\begin{table*}[!ht]
\centering
\caption{Results on VSI-Bench full split. Methods are grouped by model type. Best results among models are shown in \textbf{bold}, second best results are \underline{underlined}. ``--'' indicates results not reported. Space Tokens variants are highlighted \sethlcolor{cyan} \hl{blue} and the pretrained checkpoints are highlighted \sethlcolor{orange} \hl{orange}.}
\label{tab:vsi_bench}

\resizebox{\textwidth}{!}{
\begin{tabular}{lcccccccccc}
\toprule
\multirow{2}{*}{Method}
& \multirow{2}{*}{Avg.}
& \multicolumn{4}{c}{Numerical Questions}
& \multicolumn{4}{c}{Multiple-Choice Questions} \\
\cmidrule(lr){3-6}
\cmidrule(lr){7-10}
&
& Obj. Cnt.
& Abs. Dist.
& Obj. Size
& Room Size
& Rel. Dist.
& Rel. Dir.
& Route Plan
& Appr. Order \\
\midrule

\multicolumn{10}{l}{\textit{Proprietary Models}}\\
\midrule

GPT-4o \cite{openai2024gpt4technicalreport}                & 34.0 & 46.2 & 5.3  & 43.8 & 38.2 & 37.0 & 41.3 & 31.5 & 28.5 \\
Gemini-1.5 Flash \cite{gemini}          & 42.1 & 49.8 & 30.8 & 53.5 & 54.4 & 37.7 & 41.0 & 31.5 & 37.8 \\
Gemini-1.5 Pro             & 45.4 & 56.2 & 30.9 & 64.1 & 43.6 & 51.3 & 46.3 & 36.0 & 34.6 \\
Gemini 3 Pro               & 52.5 & 38.0 & 37.8 & 72.7 & 44.1 & 59.9 & 55.7 & \underline{45.9} & 66.0 \\
GPT-5.2                    & 54.5 & -- & -- & -- & -- & -- & -- & -- & -- \\

\midrule
\multicolumn{10}{l}{\textit{Open-Source General VLMs}}\\
\midrule

LLaVA-Next-Video-7B        & 35.6 & 48.5 & 14.0 & 47.8 & 24.2 & 43.5 & 42.4 & 34.0 & 30.6 \\
LLaVA-OneVision-72B \cite{Li_onevision}       & 40.9 & 43.5 & 23.9 & 57.6 & 37.5 & 42.5 & 39.9 & 32.5 & 44.6 \\
\rowcolor{orange!9}
InternVL3-8B \cite{Zhu_internvl} & 42.1 & 66.0 & 34.9 & 43.6 & 47.5 & 48.0 & 39.3 & 26.3 & 31.4 \\
InternVL3-78B & 48.5 & 71.2 & 53.7 & 44.4 & 39.5 & 55.9 & 39.5 & 28.9 & 54.5 \\
\rowcolor{orange!9}
Qwen3-VL-8B \cite{Bai_qwenvl3}                & 59.8 & 67.1 & 52.3 & 76.1 & 56.7 & 59.9 & 54.1 & 38.7 & 73.3 \\

\midrule
\multicolumn{10}{l}{\textit{Architecture-Customized Spatial Models}}\\
\midrule
Spatial-MLLM \cite{Wu_spatial_mllm}     & 48.4 & 65.3 & 34.8 & 63.1 & 45.1 & 41.3 & 46.2 & 33.5 & 46.3 \\
VLM-3R \cite{Fan_vlm3r}                     & 60.9 & 70.2 & 49.4 & 69.2 & 67.1 & 65.4 & 80.5 & 45.4 & 40.1 \\
SpaceMind \cite{Zhao_spacemind}        & \underline{69.6} & 73.3 & \underline{61.4} & 77.3 & 74.2 & 67.2 & \underline{88.4} & 44.3 & 70.6 \\
GeoThinker \cite{Li_thinkwithgeometry}       & \textbf{72.6} & \textbf{74.6} & \textbf{62.8} & 76.9 & 74.9 & \textbf{75.9} & \textbf{90.2} & \textbf{48.4} & 77.5 \\

\midrule
\multicolumn{10}{l}{\textit{Training-Only (No Architecture Modification)}}\\
\midrule

SpaceR \cite{Ouyang_spacer} & 45.5 & 57.8 & 28.2 & 59.9 & 47.1 & 40.1 & 45.4 & 33.5 & 52.1 \\
ViLaSR \cite{Wu_vilasr}                     & 45.4 & 63.5 & 34.4 & 60.6 & 30.9 & 48.9 & 45.2 & 30.4 & 49.2 \\

SenseNova-SI-1.2 (InternVL3-8B) \cite{Cai_sensenova}             & \underline{69.6} & 72.7 & 56.0 & 77.1 & \underline{75.7} & \underline{70.4} & 81.7 & 42.8 & \textbf{79.9} \\

\rowcolor{orange!9}
SenseNova-SI-1.3 (Qwen3-VL-8B)\textsuperscript{1} & 67.6 & 72.7 & 54.9 & 77.1 & 66.2 & 70.1 & 81.0 & 42.3 & 76.4 \\

\midrule
\multicolumn{10}{l}{Space Tokens}\\
\midrule

\rowcolor{blue!6}
InternVL3-8B & 55.9 & 68.8 & 49.7 & 74.5 & 72.7 & 55.5 & 36.5 & 26.8 & 62.9 \\
\rowcolor{blue!6}
\quad $\Delta$ \textit{Improvement} & \textcolor{blue}{$\uparrow$ 13.8} & \textcolor{blue}{$\uparrow$ 2.8} & \textcolor{blue}{$\uparrow$ 14.8} & \textcolor{blue}{$\uparrow$ 30.9} & \textcolor{blue}{$\uparrow$ 25.2} & \textcolor{blue}{$\uparrow$ 7.5} & \textcolor{red}{$\downarrow$ 2.8} & \textcolor{blue}{$\uparrow$ 0.5} & \textcolor{blue}{$\uparrow$ 31.5} \\

\rowcolor{blue!6}
Qwen3-VL-8B & 64.1 & \underline{74.4} & 57.4 & \underline{77.8} & 72.5 & 64.4 & 55.1 & 37.6 & 73.5 \\
\rowcolor{blue!6}
\quad $\Delta$ \textit{Improvement} & \textcolor{blue}{$\uparrow$ 4.3} & \textcolor{blue}{$\uparrow$ 7.3} & \textcolor{blue}{$\uparrow$ 5.1} & \textcolor{blue}{$\uparrow$ 1.7} & \textcolor{blue}{$\uparrow$ 15.8} & \textcolor{blue}{$\uparrow$ 4.5} & \textcolor{blue}{$\uparrow$ 1.0} & \textcolor{red}{$\downarrow$ 1.1} & \textcolor{blue}{$\uparrow$ 0.2} \\

\rowcolor{blue!6}
SenseNova-SI-1.3 (Qwen3-VL-8B) & 69.0 & 74.0 & 59.4 & \textbf{79.0} & \textbf{76.7} & 68.0 & 77.7 & 38.6 & \underline{78.1} \\
\rowcolor{blue!6}
\quad $\Delta$ \textit{Improvement} & \textcolor{blue}{$\uparrow$ 1.4} & \textcolor{blue}{$\uparrow$ 1.3} & \textcolor{blue}{$\uparrow$ 4.5} & \textcolor{blue}{$\uparrow$ 1.9} & \textcolor{blue}{$\uparrow$ 10.5} & \textcolor{red}{$\downarrow$ 2.1} & \textcolor{red}{$\downarrow$ 3.3} & \textcolor{red}{$\downarrow$ 3.7} & \textcolor{blue}{$\uparrow$ 1.7} \\


\bottomrule

Human               & 79.2 & 94.3 & 47.0 & 60.4 & 45.9 & 94.7 & 95.8 & 95.8 & 100 \\

\bottomrule
\end{tabular}
}
\par\smallskip
\begin{minipage}{\textwidth}
\footnotesize\raggedright
\textsuperscript{1}SenseNova team reports 67.8 overall, without reporting task results. We re-evaluated the checkpoint, noting that minor discrepancies may occur due to different dependencies and GPU hardware.
\end{minipage}

\end{table*}

\begin{table}[!ht] \centering \caption{Results on OOD benchmarks. We evaluate the trained model on related but unseen benchmarks to assess whether it preserves its general capabilities beyond the training tasks. BLINK and CV-Bench evaluate visual perception and spatial understanding, while NExT-QA measures general video-based question answering. The results show gains on BLINK and NExT-QA, with modest decreases on CV-Bench. }  \label{tab:benchmarks} 

\resizebox{\linewidth}{!}{ \begin{tabular}{lccc} \toprule Method & \textbf{BLINK} & \textbf{CV-Bench} & \textbf{NExT-QA} \\ \midrule Qwen3-VL-8B + SenseNova-SI & 64.4 & 88.4 & 74.4 \\ 
Qwen3-VL-8B + SenseNova-SI + Stage 1 \& 2 & 64.9 & 86.9 & 80.0 \\ \quad $\Delta$ \textit{Improvement} & \textcolor{blue}{$\uparrow 0.5$} & \textcolor{red}{$\downarrow 1.5$} & \textcolor{blue}{$\uparrow 5.6$} \\
Qwen3-VL-8B + SenseNova-SI + Stage 1 \& 2 \& 3 & 64.6 & 87.0 & 81.3 \\ \quad $\Delta$ \textit{Improvement} & \textcolor{blue}{$\uparrow 0.2$} & \textcolor{red}{$\downarrow 1.4$} & \textcolor{blue}{$\uparrow 6.9$} \\ \bottomrule \end{tabular} } \end{table}

\subsection{Implementation Details}
\subsubsection{Training Setup}
Space Tokens models are fine-tuned on top of the pretrained InternVL3-8B, Qwen3-VL-8B and SenseNova-SI-1.3 VLMs, applying low-rank adaption (LoRA) \cite{Hu_lora}. We use rank 16 and a scaling factor of 32 for the Qwen3-VL-based models, and rank 32 and a scaling factor of 64 for InternVL3-8B. GRPO is applied only to the Qwen3-VL-based models; before this stage, the LoRA adapter from the prior SFT stages is merged, and a new LoRA adapter with the same configuration is instantiated. A cosine learning rate scheduler is employed using a learning rate of $5 \times10^{-5}$ and $1 \times 10^{-6}$ for the SFT and GRPO stages, respectively. More detailed training configurations are described in the appendix.

\subsubsection{Data}
The Space Tokens models are fine-tuned for 1 epoch using the VICA-322K \cite{feng_vica} dataset for both stage 1 \& 2, and stage 3 exposes the model to an additional 10,000 samples. VICA-322K is chosen because it provides large-scale instructional video sequences containing diverse environments and objects, making it well-suited for learning continuous scene-level geometry and object-centric spatial relationships. During data processing, 32 images are uniformly sampled from the input video and are resized to $448 \times 448$ pixels. However, spatial reasoning is performed over the 6 most informative frames, as this configuration was found to provide the best performance by balancing relevant spatial information against redundant and noisy content. Fig~\ref{fig:prompts} outlines the prompt structure used in each stage of training. Further details on dataset configuration and processing are provided in the appendix.

\subsubsection{Benchmark}
We mainly evaluate Space Tokens on VSI-Bench \cite{Yang_vsi_bench}, which contains over 5,000 question and answer pairs designed to evaluate spatial understanding in indoor environments, with data curated from ARKitScenes \cite{Baruch_arkitscenes}, ScanNet \cite{Dai_scannet}, and ScanNet++ \cite{yeshwanth_scannetpp}. VSI-Bench considers two categories of questions, multiple choice (MCA) and numerical answer (NA), which span eight tasks: object counting, absolute distance, object size, room size, relative distance, relative direction, route planning, and object appearance order. Evaluation on VSI-Bench is conducted following the official protocol, and uses mean accuracy for MCA and relative accuracy for NA.

To verify that our training procedure does not induce catastrophic forgetting or degrade the VLM’s general capabilities, we also evaluate the resulting models on several related but out-of-distribution (OOD) benchmarks. Specifically, BLINK~\cite{fu2024blink} and CV-Bench~\cite{tong2024cambrian} assess visual perception and spatial understanding, while NExT-QA~\cite{xiao2021next} assesses general video-based question answering.

\subsection{Quantitative Results}

Table~\ref{tab:vsi_bench} reports results on the full VSI-Bench benchmark, comparing Space Tokens with proprietary VLMs, open-source general-purpose VLMs, architecture-customized spatial models, and training-only approaches. Space Tokens improves performance across two distinct VLM architectures, InternVL3 and Qwen3-VL, while preserving their underlying architectures. Starting from the original pretrained InternVL3-8B checkpoint, our method increases the overall score from 42.1 to 55.9 (+13.8 percentage points), improving seven of eight task metrics. The largest gains occur in the appearance order (+31.5), object size (+30.9), and room size (+25.2) tasks. Starting from the pretrained Qwen3-VL-8B model, Space Tokens improves the overall score from 59.8 to 64.1 (+4.3 percentage points). Applied to the spatially adapted SenseNova-SI-1.3 checkpoint, it further improves performance from 67.6 to 69.0 (+1.4 percentage points), achieving state-of-the-art performance of 79.0 on object size and 76.7 on room size estimation. These two tasks directly assess the object extent and global scene geometry, respectively, closely aligning with the object-centric 3D bounding-box and scene-level reconstruction modalities introduced by Space Tokens. These results support the method's applicability across VLM architectures and highlight its benefit on tasks most directly related to incorporated spatial representations.
As shown in Table~\ref{tab:vsi_bench}, GeoThinker~\cite{Li_thinkwithgeometry} achieves higher average performance and outperforms our model on certain question categories. However, GeoThinker integrates VGGT into its architecture, introducing approximately 1B additional parameters. In contrast, our approach improves spatial reasoning capabilities without increasing the model's parameter count, offering a more parameter-efficient solution.

\begin{figure*}[t]
    \centering

    \includegraphics[width=\linewidth]{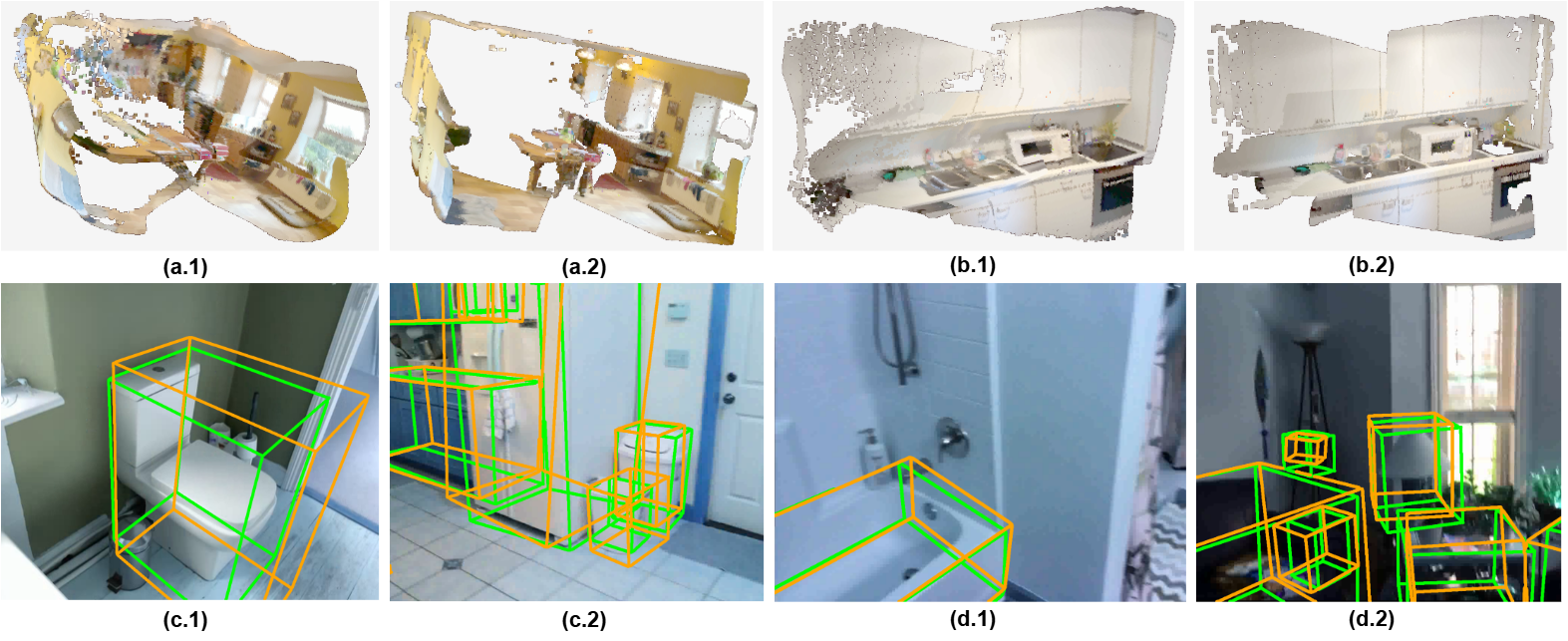}


    \caption{Qualitative comparison of generated 3D scenes (Top) and predicted 3D object bounding boxes (Bottom). (a.1) and (b.1) showcase the decoded 3D token predictions, while (a.2) and (b.2) depict the corresponding expert predictions from VGGT. In the bottom images, (c.1-4), the orange lines show the prediction and the green lines show the ground truth.}
    
    \label{fig:qualitative}
\end{figure*}

We further evaluate our model on out-of-distribution benchmarks, as reported in Table~\ref{tab:benchmarks}. Specifically, we compare the SenseNova-SI-1.3 checkpoint with our model after Stage 1 \& 2 as well as after all three training stages. On BLINK, a comprehensive benchmark for visual perception and spatial reasoning, our model achieves improvements of 0.5 and 0.2 percentage points, respectively, while exhibiting modest decreases of 1.5 and 1.4 percentage points on CV-Bench. It is worth noting that our model is trained to process 32 input images and perform reasoning over 6 selected frames, whereas BLINK and CV-Bench typically contain substantially fewer images. To accommodate examples with fewer than 6 frames, we use a balanced frame allocation strategy where available frames are repeated as necessary to preserve the expected input structure.

We further evaluate our model on NExT-QA, a general video question-answering benchmark, where it achieves substantial improvements of 5.6 and 6.9 percentage points, respectively. These results suggest that the enhanced spatial representations and reasoning capabilities may generalize beyond spatial tasks, benefiting broader video understanding.

\subsection{Qualitative Results}

To show the visual verifiability of our method, Fig.~\ref{fig:qualitative} presents several examples of reconstructed 3D scenes and predicted 3D object bounding boxes. Each modality is decoded from a dedicated subset of the VLM’s latent states using lightweight projection, cross-attention and MLP layers. Unlike specialized 3D reconstruction or object-prediction methods, which typically rely on large models trained on extensive datasets, our decoders are trained with comparatively limited supervision. We therefore do not expect perfectly accurate reconstructions. Nevertheless, these visualizations confirm that the learned latent representations capture meaningful geometric information, while our quantitative results demonstrate their effectiveness in improving VQA reasoning.

\subsection{Ablations}
\subsubsection{Data Scaling Ablations}
\begin{table}[!tp]
\centering
\caption{Ablation study on data scaling. Additional training on VICA-322K alone slightly degrades performance, whereas Space Tokens improves VSI-Bench average score after SFT and RL fine-tuning. Results show cumulative training exposure starting from SenseNova's upstream training on 14M samples.}
\label{tab:data_ablations}

\resizebox{\linewidth}{!}{%
\begin{tabular}{lcc}
\toprule
Method & Data Samples (Million) & VSI-Bench Avg. \\
\midrule
Qwen3-VL-8B & -- & 59.8 \\
Qwen3-VL-8B + SenseNova-SI & 14M & 67.6 \\
Qwen3-VL-8B + SenseNova-SI + VICA-322K & 14.6M & 67.4 \\
Qwen3-VL-8B + SenseNova-SI + stage 1 \& 2 & 14.6M & 68.6 \\
Qwen3-VL-8B + SenseNova-SI + stage 1 \& 2 \& 3 & 14.6M & \textbf{69.0} \\
\bottomrule
\end{tabular}%
}
\end{table}
To isolate the contribution of the proposed tokens, we additionally fine-tune SenseNova-SI-1.3 on the same VICA-322K dataset without introducing our Space Tokens training pipeline for an equivalent data exposure. The results are given in Table \ref{tab:data_ablations}; this baseline achieves a lower overall average score on VSI-Bench compared to the pre-trained SenseNova-SI-1.3 baseline, indicating that the model may be overfitting to the VICA-322K dataset, as this data is already present in SenseNova-SI. However, after training on the same data with Space Tokens, the results surpass the SenseNova-SI-1.3 checkpoint. This validates that the observed improvements in the other experiments arise from the proposed spatial tokens rather than additional data scaling.

\begin{table}[!tp]
    \centering
    \caption{Ablation study on the impact of specific spatial token task representations on the VSI-Bench full split.}
    \label{tab:task_ablations}

    \begin{tabular}{ccc}
    \hline
    \textbf{Box Tokens} & \textbf{3D Tokens} & \textbf{VSI-Bench Score} \\ \hline
    \checkmark        & \texttimes        & 67.8                 \\
    \texttimes        & \checkmark        & 67.4                 \\
    \checkmark        & \checkmark        & \textbf{68.4}        \\ \hline
    \end{tabular}
\end{table}

\subsubsection{Spatial Task Ablations}

We evaluate the contribution of each spatial modality by training all variants on the full VICA-322K dataset in Stage 1, followed by 40K samples in Stage 2 to reduce computational cost. As shown in Table~\ref{tab:task_ablations}, combining scene-level 3D reconstruction and object-level 3D bounding-box supervision achieves the best VSI-Bench score among these variants at 68.4, compared with 67.8 for box tokens alone and 67.4 for 3D tokens alone. The combined model exceeds the original SenseNova-SI-1.3 score of 67.6 despite the reduced Stage 2 training budget. These results support the complementary contributions of the two spatial modalities.




\section{Conclusion}

We introduce Space Tokens, a lightweight framework that integrates scene-level geometry and object-centric spatial information into VLMs through continuous latent tokens. Our method enables the model to learn rich spatial priors without architectural modifications. Unlike approaches that only evaluate spatial understanding through downstream performance, Space Tokens produces verifiable representations that encode meaningful geometric information and support reasoning. Experiments demonstrate improved overall VSI-Bench scores across the tested backbones, including state-of-the-art performance on geometry-intensive tasks such as object- and room-size estimation. OOD evaluation shows gains on BLINK and NExT-QA, with modest decreases on CV-Bench. Ablations further show that these improvements arise from the proposed spatial representations rather than additional data scaling. More broadly, Space Tokens provides a general interface for incorporating interpretable continuous modalities into chain-of-thought reasoning beyond 3D geometry.

\section*{LLM Usage Statement}
OpenAI Codex was used to assist with the implementation and debugging of selected components of the proposed framework, as well as code refactoring and development support. All AI-assisted code was reviewed and validated by the authors. Codex also assisted with migrating the manuscript to the CVPR 2027 \LaTeX{} template, including document setup, formatting, and compilation checks.

\FloatBarrier
{
    \small
    \bibliographystyle{ieeenat_fullname}
    \bibliography{references}

@String(CVPR= {IEEE Conf. Comput. Vis. Pattern Recog.})

@String(CVPR  = {CVPR})

@InProceedings{Yang_vsi_bench,
    author    = {Yang, Jihan and Yang, Shusheng and Gupta, Anjali W. and Han, Rilyn and Fei-Fei, Li and Xie, Saining},
    title     = {Thinking in Space: How Multimodal Large Language Models See, Remember, and Recall Spaces},
    booktitle = {Proceedings of the IEEE/CVF Conference on Computer Vision and Pattern Recognition (CVPR)},
    month     = {June},
    year      = {2025},
    pages     = {10632-10643}
}

@misc{feng_vica,
      title={Visuospatial Cognitive Assistant}, 
      author={Qi Feng},
      year={2025},
      eprint={2505.12312},
      archivePrefix={arXiv},
      primaryClass={cs.CV},
      url={https://arxiv.org/abs/2505.12312}, 
}

@inproceedings{Baruch_arkitscenes,
 author = {Dehghan, Afshin and Baruch, Gilad and Chen, Zhuoyuan and Feigin, Yuri and Fu, Peter and Gebauer, Thomas and Kurz, Daniel and Dimry, Tal and Joffe, Brandon and Schwartz, Arik and Shulman, Elad},
 booktitle = {Proceedings of the Neural Information Processing Systems Track on Datasets and Benchmarks},
 editor = {J. Vanschoren and S. Yeung},
 pages = {},
 title = {ARKitScenes: A Diverse Real-World Dataset For 3D Indoor Scene Understanding Using Mobile RGB-D Data},
 url = {https://datasets-benchmarks-proceedings.neurips.cc/paper_files/paper/2021/file/66f041e16a60928b05a7e228a89c3799-Paper-round1.pdf},
 volume = {1},
 year = {2021}
}

@inproceedings{Dai_scannet,
    title={ScanNet: Richly-annotated 3D Reconstructions of Indoor Scenes},
    author={Dai, Angela and Chang, Angel X. and Savva, Manolis and Halber, Maciej and Funkhouser, Thomas and Nie{\ss}ner, Matthias},
    booktitle = {Proc. Computer Vision and Pattern Recognition (CVPR), IEEE},
    year = {2017}
}

@inproceedings{yeshwanth_scannetpp,
  title={Scannet++: A high-fidelity dataset of 3d indoor scenes},
  author={Yeshwanth, Chandan and Liu, Yueh-Cheng and Nie{\ss}ner, Matthias and Dai, Angela},
  booktitle={Proceedings of the IEEE/CVF International Conference on Computer Vision},
  pages={12--22},
  year={2023}
}

@misc{Zhang_ablate_to_validate,
      title={Ablate-to-Validate: Are Vision-Language Models Really Using Continuous Thought Tokens?}, 
      author={Tianyi Zhang and Mahtab Bigverdi and Ranjay Krishna},
      year={2026},
      eprint={2605.21642},
      archivePrefix={arXiv},
      primaryClass={cs.CV},
      url={https://arxiv.org/abs/2605.21642}, 
}

@misc{Shao_GRPO,
      title={DeepSeekMath: Pushing the Limits of Mathematical Reasoning in Open Language Models}, 
      author={Zhihong Shao and Peiyi Wang and Qihao Zhu and Runxin Xu and Junxiao Song and Xiao Bi and Haowei Zhang and Mingchuan Zhang and Y. K. Li and Y. Wu and Daya Guo},
      year={2024},
      eprint={2402.03300},
      archivePrefix={arXiv},
      primaryClass={cs.CL},
      url={https://arxiv.org/abs/2402.03300}, 
}

@misc{Hu_lora,
      title={LoRA: Low-Rank Adaptation of Large Language Models}, 
      author={Edward J. Hu and Yelong Shen and Phillip Wallis and Zeyuan Allen-Zhu and Yuanzhi Li and Shean Wang and Lu Wang and Weizhu Chen},
      year={2021},
      eprint={2106.09685},
      archivePrefix={arXiv},
      primaryClass={cs.CL},
      url={https://arxiv.org/abs/2106.09685}, 
}

@misc{Li_onevision,
      title={LLaVA-OneVision: Easy Visual Task Transfer}, 
      author={Bo Li and Yuanhan Zhang and Dong Guo and Renrui Zhang and Feng Li and Hao Zhang and Kaichen Zhang and Peiyuan Zhang and Yanwei Li and Ziwei Liu and Chunyuan Li},
      year={2024},
      eprint={2408.03326},
      archivePrefix={arXiv},
      primaryClass={cs.CV},
      url={https://arxiv.org/abs/2408.03326}, 
}

@article{Ouyang_spacer,
  title={SpaceR: Reinforcing MLLMs in Video Spatial Reasoning},
  author={Ouyang, Kun and Liu, Yuanxin and Wu, Haoning and Liu, Yi and Zhou, Hao and Zhou, Jie and Meng, Fandong and Sun, Xu},
  journal={arXiv preprint arXiv:2504.01805},
  year={2025}
}

@misc{Wu_vilasr,
      title={Reinforcing Spatial Reasoning in Vision-Language Models with Interwoven Thinking and Visual Drawing}, 
      author={Junfei Wu and Jian Guan and Kaituo Feng and Qiang Liu and Shu Wu and Liang Wang and Wei Wu and Tieniu Tan},
      year={2025},
      eprint={2506.09965},
      archivePrefix={arXiv},
      primaryClass={cs.CV},
      url={https://arxiv.org/abs/2506.09965}, 
}

@inproceedings{Wu_spatial_mllm,
 author = {Wu, Diankun and Liu, Fangfu and Hung, Yi-Hsin and Duan, Yueqi},
 booktitle = {Advances in Neural Information Processing Systems},
 editor = {D. Belgrave and C. Zhang and H. Lin and R. Pascanu and P. Koniusz and M. Ghassemi and N. Chen},
 pages = {13569--13597},
 publisher = {Curran Associates, Inc.},
 title = {Spatial-MLLM: Boosting MLLM Capabilities in Visual-based Spatial Intelligence},
 url = {https://proceedings.neurips.cc/paper_files/paper/2025/file/142b166c8b18b43774c4f0e0c1e9948e-Paper-Conference.pdf},
 volume = {38},
 year = {2025}
}

@misc{Zhao_spacemind,
      title={SpaceMind: Camera-Guided Modality Fusion for Spatial Reasoning in Vision-Language Models}, 
      author={Ruosen Zhao and Zhikang Zhang and Jialei Xu and Jiahao Chang and Dong Chen and Lingyun Li and Weijian Sun and Zizhuang Wei},
      year={2025},
      eprint={2511.23075},
      archivePrefix={arXiv},
      primaryClass={cs.CV},
      url={https://arxiv.org/abs/2511.23075}, 
}

@misc{Li_thinkwithgeometry,
      title={Thinking with Geometry: Active Geometry Integration for Spatial Reasoning}, 
      author={Haoyuan Li and Qihang Cao and Tao Tang and Kun Xiang and Zihan Guo and Jianhua Han and Hang Xu and Xiaodan Liang},
      year={2026},
      eprint={2602.06037},
      archivePrefix={arXiv},
      primaryClass={cs.CV},
      url={https://arxiv.org/abs/2602.06037}, 
}

@misc{Zheng_vgllm,
      title={Learning from Videos for 3D World: Enhancing MLLMs with 3D Vision Geometry Priors}, 
      author={Duo Zheng and Shijia Huang and Yanyang Li and Liwei Wang},
      year={2025},
      eprint={2505.24625},
      archivePrefix={arXiv},
      primaryClass={cs.CV},
      url={https://arxiv.org/abs/2505.24625}, 
}

@InProceedings{Yang_mirage,
    author    = {Yang, Zeyuan and Yu, Xueyang and Chen, Delin and Shen, Maohao and Gan, Chuang},
    title     = {Machine Mental Imagery: Empower Multimodal Reasoning with Latent Visual Tokens},
    booktitle = {Proceedings of the IEEE/CVF Conference on Computer Vision and Pattern Recognition (CVPR)},
    month     = {June},
    year      = {2026},
    pages     = {33510-33520}
}

@InProceedings{Ray_mull,
    author    = {Ray, Arijit and Abdelkader, Ahmed and Mao, Chengzhi and Plummer, Bryan A. and Saenko, Kate and Krishna, Ranjay and Guibas, Leonidas and Chu, Wen-Sheng},
    title     = {Mull-Tokens:  Modality-Agnostic Latent Thinking},
    booktitle = {Proceedings of the IEEE/CVF Conference on Computer Vision and Pattern Recognition (CVPR) Findings},
    month     = {June},
    year      = {2026},
    pages     = {9477-9488}
}

@article{Qin_covt,
  title={Chain-of-Visual-Thought: Teaching VLMs to See and Think Better with Continuous Visual Tokens},
  author={Qin, Yiming and Wei, Bomin and Ge, Jiaxin and Kallidromitis, Konstantinos and Fu, Stephanie and Darrell, Trevor and Wang, Xudong},
  journal={arXiv preprint arXiv:2511.19418},
  year={2025}
}

@misc{Chen_thinkwith3d,
      title={Think with 3D: Geometric Imagination Grounded Spatial Reasoning from Limited Views}, 
      author={Zhangquan Chen and Manyuan Zhang and Xinlei Yu and Xufang Luo and Mingze Sun and Zihao Pan and Xiang An and Yan Feng and Peng Pei and Xunliang Cai and Ruqi Huang},
      year={2026},
      eprint={2510.18632},
      archivePrefix={arXiv},
      primaryClass={cs.CV},
      url={https://arxiv.org/abs/2510.18632}, 
}

@misc{Cai_sensenova,
      title={Scaling Spatial Intelligence with Multimodal Foundation Models}, 
      author={Zhongang Cai and Ruisi Wang and Chenyang Gu and Fanyi Pu and Junxiang Xu and Yubo Wang and Wanqi Yin and Zhitao Yang and Chen Wei and Qingping Sun and Tongxi Zhou and Jiaqi Li and Hui En Pang and Oscar Qian and Yukun Wei and Zhiqian Lin and Xuanke Shi and Kewang Deng and Xiaoyang Han and Zukai Chen and Xiangyu Fan and Hanming Deng and Lewei Lu and Liang Pan and Bo Li and Ziwei Liu and Quan Wang and Dahua Lin and Lei Yang},
      year={2026},
      eprint={2511.13719},
      archivePrefix={arXiv},
      primaryClass={cs.CV},
      url={https://arxiv.org/abs/2511.13719}, 
}

@misc{Cai_EASI,
      title={Holistic Evaluation of Multimodal LLMs on Spatial Intelligence}, 
      author={Zhongang Cai and Yubo Wang and Qingping Sun and Ruisi Wang and Chenyang Gu and Wanqi Yin and Zhiqian Lin and Zhitao Yang and Chen Wei and Oscar Qian and Hui En Pang and Xuanke Shi and Kewang Deng and Xiaoyang Han and Zukai Chen and Jiaqi Li and Xiangyu Fan and Hanming Deng and Lewei Lu and Bo Li and Ziwei Liu and Quan Wang and Dahua Lin and Lei Yang},
      year={2025},
      eprint={2508.13142},
      archivePrefix={arXiv},
      primaryClass={cs.CV},
      url={https://arxiv.org/abs/2508.13142}, 
}

@misc{Fan_vlm3r,
      title={VLM-3R: Vision-Language Models Augmented with Instruction-Aligned 3D Reconstruction}, 
      author={Zhiwen Fan and Jian Zhang and Renjie Li and Junge Zhang and Runjin Chen and Hezhen Hu and Kevin Wang and Huaizhi Qu and Shijie Zhou and Dilin Wang and Zhicheng Yan and Hongyu Xu and Justin Theiss and Tianlong Chen and Jiachen Li and Zhengzhong Tu and Zhangyang Wang and Rakesh Ranjan},
      year={2026},
      eprint={2505.20279},
      archivePrefix={arXiv},
      primaryClass={cs.CV},
      url={https://arxiv.org/abs/2505.20279}, 
}

@INPROCEEDINGS{Wang_VGGT,
  author={Wang, Jianyuan and Chen, Minghao and Karaev, Nikita and Vedaldi, Andrea and Rupprecht, Christian and Novotny, David},
  booktitle={2025 IEEE/CVF Conference on Computer Vision and Pattern Recognition (CVPR)}, 
  title={VGGT: Visual Geometry Grounded Transformer}, 
  year={2025},
  volume={},
  number={},
  pages={5294-5306},
  doi={10.1109/CVPR52734.2025.00499}}

@misc{Bai_qwenvl3,
      title={Qwen3-VL Technical Report}, 
      author={{Qwen Team}},
      year={2025},
      eprint={2511.21631},
      archivePrefix={arXiv},
      primaryClass={cs.CV},
      url={https://arxiv.org/abs/2511.21631}, 
}

@InProceedings{Li_blip2,
  title = 	 {{BLIP}-2: Bootstrapping Language-Image Pre-training with Frozen Image Encoders and Large Language Models},
  author =       {Li, Junnan and Li, Dongxu and Savarese, Silvio and Hoi, Steven},
  booktitle = 	 {Proceedings of the 40th International Conference on Machine Learning},
  pages = 	 {19730--19742},
  year = 	 {2023},
  editor = 	 {Krause, Andreas and Brunskill, Emma and Cho, Kyunghyun and Engelhardt, Barbara and Sabato, Sivan and Scarlett, Jonathan},
  volume = 	 {202},
  series = 	 {Proceedings of Machine Learning Research},
  month = 	 {23--29 Jul},
  publisher =    {PMLR},
  url = 	 {https://proceedings.mlr.press/v202/li23q.html}
}

@misc{Deng_bagel,
      title={Emerging Properties in Unified Multimodal Pretraining}, 
      author={Chaorui Deng and Deyao Zhu and Kunchang Li and Chenhui Gou and Feng Li and Zeyu Wang and Shu Zhong and Weihao Yu and Xiaonan Nie and Ziang Song and Guang Shi and Haoqi Fan},
      year={2025},
      eprint={2505.14683},
      archivePrefix={arXiv},
      primaryClass={cs.CV},
      url={https://arxiv.org/abs/2505.14683}, 
}

@misc{openai2024gpt4technicalreport,
      title={GPT-4 Technical Report}, 
      author={{GPT Team}},
      year={2024},
      eprint={2303.08774},
      archivePrefix={arXiv},
      primaryClass={cs.CL},
      url={https://arxiv.org/abs/2303.08774}, 
}

@misc{Zhu_internvl,
      title={InternVL3: Exploring Advanced Training and Test-Time Recipes for Open-Source Multimodal Models}, 
      author={{InternVL Team}},
      year={2025},
      eprint={2504.10479},
      archivePrefix={arXiv},
      primaryClass={cs.CV},
      url={https://arxiv.org/abs/2504.10479}, 
}

@misc{gemini,
      title={Gemini: A Family of Highly Capable Multimodal Models}, 
      author={{Gemini Team}},
      year={2025},
      eprint={2312.11805},
      archivePrefix={arXiv},
      primaryClass={cs.CL},
      url={https://arxiv.org/abs/2312.11805}, 
}

@misc{gemma,
      title={Gemma 3 Technical Report}, 
      author={{Gemma Team}},
      year={2025},
      eprint={2503.19786},
      archivePrefix={arXiv},
      primaryClass={cs.CL},
      url={https://arxiv.org/abs/2503.19786}, 
}

@article{wang2025n3d,
  title={N3D-VLM: Native 3D Grounding Enables Accurate Spatial Reasoning in Vision-Language Models},
  author={Wang, Yuxin and Ke, Lei and Zhang, Boqiang and Qu, Tianyuan and Yu, Hanxun and Huang, Zhenpeng and Yu, Meng and Xu, Dan and Yu, Dong},
  journal={arXiv preprint arXiv:2512.16561},
  year={2025}
}

@article{man2025locateanything3d,
  title={LocateAnything3D: Vision-Language 3D Detection with Chain-of-Sight},
  author={Man, Yunze and Wang, Shihao and Zhang, Guowen and Bjorck, Johan and Li, Zhiqi and Gui, Liang-Yan and Fan, Jim and Kautz, Jan and Wang, Yu-Xiong and Yu, Zhiding},
  journal={arXiv preprint arXiv:2511.20648},
  year={2025}
}

@article{liu2025abstract,
  title={Abstract 3D Perception for Spatial Intelligence in Vision-Language Models},
  author={Liu, Yifan and Zhan, Fangneng and Zhou, Kaichen and Du, Yilun and Liang, Paul Pu and Pfister, Hanspeter},
  journal={arXiv preprint arXiv:2511.10946},
  year={2025}
}

@article{zhang2026ssr,
  title={SSR: Pushing the Limit of Spatial Intelligence with Structured Scene Reasoning},
  author={Zhang, Yi and Xia, Youya and Wang, Yong and Song, Meng and Wu, Xin and Wan, Wenjun and Liu, Bingbing and Ye, AiXue and Zhang, Hongbo and Wen, Feng},
  journal={arXiv preprint arXiv:2603.00409},
  year={2026}
}

@article{chen2026thinking,
  title={Thinking with Spatial Code for Physical-World Video Reasoning},
  author={Chen, Jieneng and Ma, Wenxin and Yuan, Ruisheng and Zhang, Yunzhi and Wu, Jiajun and Yuille, Alan},
  journal={arXiv preprint arXiv:2603.05591},
  year={2026}
}

@inproceedings{fu2024blink,
  title={Blink: Multimodal large language models can see but not perceive},
  author={Fu, Xingyu and Hu, Yushi and Li, Bangzheng and Feng, Yu and Wang, Haoyu and Lin, Xudong and Roth, Dan and Smith, Noah A and Ma, Wei-Chiu and Krishna, Ranjay},
  booktitle={European Conference on Computer Vision},
  pages={148--166},
  year={2024},
  organization={Springer}
}

@inproceedings{xiao2021next,
  title={Next-qa: Next phase of question-answering to explaining temporal actions},
  author={Xiao, Junbin and Shang, Xindi and Yao, Angela and Chua, Tat-Seng},
  booktitle={Proceedings of the IEEE/CVF conference on computer vision and pattern recognition},
  pages={9777--9786},
  year={2021}
}

@inproceedings{tong2024cambrian,
  title={Cambrian-1: A fully open, vision-centric exploration of multimodal llms},
  author={Tong, Shengbang and Brown II, Ellis L and Wu, Penghao and Woo, Sanghyun and Iyer, Adithya Jairam and Akula, Sai Charitha and Yang, Shusheng and Yang, Jihan and Middepogu, Manoj and Wang, Ziteng and others},
  booktitle={The Thirty-eighth Annual Conference on Neural Information Processing Systems},
  year={2024}
}
}

\clearpage
\AddToHookNext{cmd/appendix/after}{%
    \renewcommand{\theHequation}{supp.\arabic{equation}}%
    \renewcommand{\theHfigure}{supp.\arabic{figure}}%
    \renewcommand{\theHtable}{supp.\arabic{table}}%
}
\appendix

\setcounter{section}{0}
\renewcommand{\thesection}{\Alph{section}}

\setcounter{equation}{0}
\renewcommand{\theequation}{S\arabic{equation}}

\setcounter{figure}{0}
\renewcommand{\thefigure}{S\arabic{figure}}

\setcounter{table}{0}
\renewcommand{\thetable}{S\arabic{table}}

\newcommand{\appsection}[1]{%
  \refstepcounter{section}
  \section*{\thesection. #1}
}

\newcommand{\appsubsection}[1]{%
  \refstepcounter{subsection}
  \subsection*{\thesubsection. #1}%
}

\newcommand{\appsubsubsection}[1]{%
  \refstepcounter{subsubsection}
  \subsubsection*{\thesection.\thesubsection.\thesubsubsection\quad #1}%
}

\maketitlesupplementary

\appsection{Methodological Details}
\appsubsection{Detailed 3D Reconstruction Formulation}
\label{app:3d_reconstruction_method}

Complementing the feature-similarity objective introduced in Eq. 2, we apply geometric reconstruction objectives to ensure learned spatial representations remain predictive of explicit 3D scene structure. Specifically, the hidden states associated with the 3D representation tokens are decoded into camera parameters, depth maps, and dense point maps using the corresponding frozen VGGT prediction heads. These outputs are supervised against VGGT's geometric estimates for the same input frames. While the similarity objective encourages the 3D tokens to capture the teacher's internal geometric representation, these reconstruction objectives constrain the preservation of information needed to recover explicit scene geometry.

The VGGT camera head estimates a camera encoding for each supervised frame that corresponds to a group of 3D tokens. It is comprised of translation, rotation and field of view components. Applying the frozen VGGT teacher head to the reconstructed representation, $z$, produces the corresponding student prediction. We supervise these predictions using an $\ell_1$ objective. Since the camera head refines predictions iteratively, we apply supervision at each refinement step

\begin{equation}
    \mathcal{L}_{\mathrm{cam}} = \frac{1}{K} \sum_{k=1}^K \gamma^{K-k}(\ell_{\mathrm{trans}}^{(k)} + \ell_{\mathrm{rot}}^{(k)} + 0.5 \ell_{\mathrm{fov}}^{(k)})
\end{equation}
where $K = 4$ refinement iterations, and each $\ell^{(k)}$ denotes the mean absolute difference between the student and teacher camera parameters at step $k$. We use $\gamma = 0.6$ to give a larger weight to later refinements.

The teacher depth map, $D_i$, is similarly obtained by applying VGGT's depth head to its aggregated features. Then, the same frozen depth head decodes the reconstructed latent features into a predicted depth map, $\hat{D}_i$, and a positive confidence map. The depth objective combines three terms, a regression objective, confidence objective and gradient consistency

\begin{equation}
    \mathcal{L}_{\mathrm{dpt}} = \mathcal{L}_{\mathrm{reg}}^{D} + \mathcal{L}_{\mathrm{conf}}^D + \mathcal{L}_{\mathrm{grad}}^D.
\end{equation}
The regression term penalizes absolute discrepancies between the student and teacher depth maps. The confidence term incorporates the predicted confidence together with logarithmic regularization to prevent confidence to be reduced without penalty. The gradient consistency term investigates differences in horizontal and vertical depth at four different spatial scales, encouraging the reconstructed representation to preserve both depth boundaries and surface structure.

Under the same procedure, VGGT's point head directly predicts a dense point map, $P_i$, where each $P_i(u, v) \in \mathbb{R}^3$ represents the 3D location of an associated image pixel, $(u, v)$. Applying this head to the reconstructed representation yields $\hat{P}_i$ and a confidence map. The point map reconstruction objective comprises three terms, a regression objective, a confidence objective, and a surface normal objective for preserving local surface geometry

\begin{equation}
    \mathcal{L}_{\mathrm{pt}} = \mathcal{L}_{\mathrm{reg}}^P + \mathcal{L}_{\mathrm{conf}}^P + \mathcal{L}_{\mathrm{normal}}^P.
\end{equation}

We set the balancing coefficients in Eq. 3 to $\lambda_{\mathrm{sim}} = \lambda_{\mathrm{cam}} = \lambda_{\mathrm{dpt}} = \lambda_{\mathrm{pt}} = 1.0$, jointly optimizing feature similarity and geometric reconstruction without additional relative scaling between these objectives. Equal weighting performed better than using the original weight that VGGT used, where $\lambda_{\mathrm{cam}} = 5.0$ while all other components remained at $1.0$ weight. Computational constraints prevented a broader hyperparameter sweep, so this configuration represents the best-performing setting among tested configurations.

\appsubsection{Detailed Object 3D Bounding Box Generation}
\label{app:bounding_box_method}

Given the LLM hidden states associated with the bounding-box tokens, denoted by $H_v$, we project them into a (d)-dimensional embedding space and normalize the resulting features along the feature dimension:

\begin{equation}
\bar{e}_v = \frac{
H_v W_{\mathrm{proj}}^\top
}{
\left\lVert H_v W_{\mathrm{proj}}^\top \right\rVert_2
}
\in \mathbb{R}^{T \times d}
\end{equation}

We subsequently apply cross-attention between a set of learnable object queries, $Q$, and the normalized token representations: 

\begin{equation}
a_v
=
\operatorname{MHA}_{\text{cross-attn}}
\left(
Q,
\bar{e}_v,
\bar{e}_v
\right),
\qquad
a_v \in \mathbb{R}^{T \times d}
\end{equation}

The $Q$ is identical for every view and is a fixed bank of learned slot queries, each notionally responsible for one candidate object. Next, the attended token features are augmented with embeddings of the normalized 2D object bounding boxes:
\begin{equation}\label{eq:bb+}
s_v
=
a_v
+
\lambda_{\mathrm{bb}} B_v W_{\mathrm{bbox}}^\top,
\qquad
\lambda_{\mathrm{bb}} = 0.2
\end{equation}

Here, $B_v$ denotes the normalized 2D bounding-box coordinates, and $W_{\mathrm{bbox}}$ projects them into the same feature space as $a_v$. This residual conditioning provides explicit object-localization cues for 3D bounding-box prediction.
To further incorporate fine-grained visual information, the visual features $X_v$ produced by the VLM’s vision backbone are first projected into the decoder embedding space and then normalized:
\begin{equation}
\tilde{X}_v = X_v W_{\mathrm{vis}}^\top
\end{equation}
\begin{equation}
\bar{X}_v = \frac{\tilde{X}_v}
{\lVert \tilde{X}_v \rVert_2}
\end{equation}

We use the object-conditioned representations $s_v$ as queries in a second cross-attention operation over the visual features:

\begin{equation}\label{eq:vis3}
g_v
=
\operatorname{MHA}_{\text{slot-vis-attn}}
\left(
s_v,
\bar{X}_v,
\tilde{X}_v
\right)
\end{equation}

The resulting visual features $g_v$ are concatenated with the object-conditioned token features $s_v$, followed by layer normalization, linear projection, and a GELU activation:

\begin{equation}\label{eq:combine}
f_v
=
\operatorname{GELU}
\left(
\operatorname{LN}
\left(
[s_v \,\Vert\, g_v]
\right)
W_2^\top
\right)
\in \mathbb{R}^{T \times 1024},
\end{equation}

Finally, a linear prediction head maps the fused representations to the 3D bounding-box parameters:
\begin{equation}
z_v
=
f_v W_{\mathrm{head}}^\top
\in \mathbb{R}^{T \times 12}
\end{equation}

We represent each predicted 3D bounding box as $\mathbf{p}_{vi}\in\mathbb{R}^{12}$, containing its camera-space center, dimensions, and 6D rotation representation. For view $v$, the $S$ prediction slots are matched to $K_v$ ground-truth boxes using Hungarian assignment:
\begin{equation}
\pi_v^{*}
=
\arg\min_{\pi\in\mathfrak{S}_S}
\sum_{i=1}^{S} C_{vi,\pi(i)},
\end{equation}
where
\begin{equation}
C_{vij}
=
\begin{cases}
\dfrac{1}{12}
\left\lVert \mathbf{p}_{vi}-\mathbf{g}_{vj}\right\rVert_1,
& j\leq K_v, \\[6pt]
\dfrac{1}{12}
\left\lVert \mathbf{p}_{vi}\right\rVert_1,
& j>K_v.
\end{cases}
\end{equation}
Columns $j>K_v$ correspond to background targets. After matching, we use
the Smooth $\ell_1$ penalty
\begin{equation}
\rho_{B}(r)
=
\begin{cases}
\dfrac{r^2}{2B}, & |r|<B, \\[6pt]
|r|-\dfrac{B}{2}, & |r|\geq B,
\end{cases}
\qquad B=0.1.
\end{equation}
The loss for prediction slot $i$ in view $v$ is
\begin{equation}
\ell_{vi}
=
w_{vi}\frac{1}{12}
\sum_{d=1}^{12}
\rho_{0.1}\!\left(p_{vid}-q_{vid}\right),
\end{equation}
where the target and weight are defined as
\begin{equation}
(\mathbf{q}_{vi},w_{vi})
=
\begin{cases}
\left(\mathbf{g}_{v,\pi_v^{*}(i)},\,1.0\right),
& \pi_v^{*}(i)\leq K_v, \\[4pt]
\left(\mathbf{0},\,0.35\right),
& \pi_v^{*}(i)>K_v.
\end{cases}
\end{equation}
The final 3D bounding-box loss is summed over prediction slots and averaged
over annotated views:
\begin{equation}
\mathcal{L}_{\mathrm{3D\text{-}BB}}
=
\frac{
\displaystyle\sum_v s_w\sum_{i=1}^{S}\ell_{vi}
}{
\displaystyle\max\!\left(1,\sum_w s_w\right)
},
\end{equation}
where $s_w\in\{0,1\}$ indicates whether view $w$ has 3D bounding-box
supervision.

\appsubsection{GRPO Refinement and Reward Design}
\label{app:reward_design}


In stage 3, we refine the stage 2 model using group relative policy optimization. We retain the spatial reasoning prompt structure used during supervised training. That is, spatial-token blocks corresponding to the selected frames are supplied as conditioning context, and the model only generates the final answer. Consequently, reinforcement learning optimizes answer quality without requiring autoregressive generation of the reasoning template. The stage 2 policy provides both the initialization and the frozen reference policy used for KL regularization.

We construct a 10,000 sample subset from VICA's original annotation covering six VSI-related task families: object counting, absolute distance, object size, room size, relative distance, and appearance order. Selection balances task families, stratifies numerical answers by quantile and multiple choice answers by option letter, and prioritizes scene and question-template diversity. Duplicate questions, conflicting labels, and unsupported answer formats are removed, and scenes appearing in the VSI-Bench evaluation set are excluded. Selection uses annotation metadata and a fixed random seed of 42.

For each input, we sample a group of $G = 8$ answer sequences with temperature $1.0$. Rewards are determined by the annotated question type. For multiple-choice questions, the predicted and reference option letters are extracted and normalized; a correct choice receives a reward of $1$ while incorrect or unparseable responses receive $0$. For numerical questions, we employ the mean relative accuracy \cite{Yang_vsi_bench} as a reward. Specifically, for a set of confidence intervals $\mathcal{C} = \{0.5, 0.55, \ldots, 0.95 \}$ the reward for prediction $\hat{y}$ and ground truth answer $y$ is given as:




\begin{table*}[t]
    \centering
    \caption{Training configurations for Stages 1 and 2 and the
    VQA-only control. All configurations use an effective batch size of 32 and LoRA learning rate of $5\times10^{-5}$. Spatial-module learning rates apply only when these modules are trainable; ``---'' indicates no update. Budgets specify scheduled training exposure, which may exceed the exposure of the checkpoint selected for reporting.}
    \label{tab:hyperparams_st12}
    \small
    \setlength{\tabcolsep}{4pt}
    \renewcommand{\arraystretch}{1.2}

    \begin{tabular}{@{}
        p{0.28\textwidth}
        c c c c c 
        p{0.17\textwidth}
        @{}}
        \toprule
        \multirow{2}{*}{Configuration}
        & \multirow{2}{*}{Stage}
        & \multicolumn{2}{c}{LoRA}
        & \multirow{2}{*}{\shortstack{Spatial-module\\LR}}
        & \multirow{2}{*}{\shortstack{Spatial\\modules}}
        & \multirow{2}{*}{VICA training budget} \\
        \cmidrule(lr){3-4}
        & & Rank & Alpha & & & \\
        \midrule

        SenseNova-SI-1.3\newline
        VQA-only control
        & Control
        & 16
        & 32
        & ---
        & Absent
        & Two epochs \\
        \midrule

        \multirow{2}{0.28\textwidth}{%
            \raggedright Qwen3-VL-8B\newline
            + Space Tokens}
        & 1
        & \multirow{2}{*}{16}
        & \multirow{2}{*}{32}
        & $1\times10^{-5}$
        & Trainable
        & \multirow{2}{0.17\textwidth}{%
            \raggedright One epoch per stage} \\
        & 2 & & & --- & Frozen & \\
        \midrule

        \multirow{2}{0.28\textwidth}{%
            \raggedright SenseNova-SI-1.3 (Qwen3-VL-8B)\newline
            + Space Tokens}
        & 1
        & \multirow{2}{*}{16}
        & \multirow{2}{*}{32}
        & $1\times10^{-5}$
        & Trainable
        & \multirow{2}{0.17\textwidth}{%
            \raggedright One epoch per stage} \\
        & 2 & & & --- & Frozen & \\
        \midrule

        \multirow{2}{0.28\textwidth}{%
            \raggedright InternVL3\newline
            + Space Tokens}
        & 1
        & \multirow{2}{*}{32}
        & \multirow{2}{*}{64}
        & $1\times10^{-5}$
        & Trainable
        & \multirow{2}{0.17\textwidth}{%
            \raggedright One epoch per stage} \\
        & 2 & & & --- & Frozen & \\
        \midrule

        \multirow{2}{0.28\textwidth}{%
            \raggedright Spatial-token ablations\newline
            (boxes / 3D / both)}
        & 1
        & \multirow{2}{*}{16}
        & \multirow{2}{*}{32}
        & $1\times10^{-5}$
        & Trainable
        & One epoch \\
        & 2
        & & &
        ---
        & Frozen
        & 40K samples \\
        \bottomrule
    \end{tabular}
\end{table*}

\begin{equation}
     R_{MRA}(\hat{y}, y) = \frac{1}{10} \sum_{\theta \in \mathcal{C}} \mathds{1} \left( \frac{\left| \hat{y} - y \right|}{y} < 1-\theta \right) .
\end{equation}
This reward provides partial credit across ten relative-error tolerances. Numerical predictions must contain a single finite value, optionally followed by units where applicable. When the reference value is zero, reward is assigned only to an exact-zero prediction. 

For rewards $\{ R_1, \ldots , R_G \}$ each associated with the same input, we compute the sequence-level advantage

\begin{equation}
    A_i = \frac{R_i - \bar{R}}{\mathrm{std}(R_1, \ldots, R_G) + \hat{\epsilon}}
\end{equation}
with $\hat{\epsilon} = 10^{-4}$. This advantage is applied to the generated answer tokens using the clipped objective in Eq. 5, with clipping width $\epsilon = 0.2$ and KL regularization coefficient $\beta = 0.04$. Old policy likelihoods are recorded before the policy update while the SFT reference remains fixed. The objective is normalized by completion length and excludes prompt tokens and positions following the first end-of-sequence token. During this stage, the spatial decoders remain frozen and auxiliary reconstruction losses are disabled. The LoRA parameters are optimized through the answer-generation objective, refining how the model uses the learned spatial representations for downstream reasoning.

\appsection{Implementation Details}
\appsubsection{Model Training}
\label{app:training_setup}

Table ~\ref{tab:hyperparams_st12} summarizes the training configurations used for stages 1 and 2 as well as the spatial-token-free control experiment. These configurations include backbone initialization, LoRA configuration, learning rate and training budget. During stage 1, all spatial modules are jointly optimized with the LoRA adapters. In stage 2, spatial modules are frozen while LoRA adaption continues. The VQA-only control uses two epochs of VICA data without spatial tokens or auxiliary objectives, matching the VICA exposure of the two-stage Space Tokens configurations.  



\begin{table}[t]
    \centering
    \caption{Stage 3 GRPO training hyperparameters (only applied to Qwen3-VL-8B models). Effective batch size counts unique prompts; each prompt produces eight sampled completions.}
    \label{tab:grpo_hyperparameters}
    \small
    \begin{tabular}{@{}lc@{}}
        \toprule
        Hyperparameter & Value \\
        \midrule
        LoRA rank & 16 \\
        LoRA alpha & 32 \\
        LoRA dropout & 0.01 \\
        Optimizer & AdamW \\
        Learning rate & $1\times10^{-6}$ \\
        Learning-rate schedule & Cosine \\
        Adam $(\beta_1,\beta_2)$ & $(0.9,0.999)$ \\
        Warmup fraction & 0.1 \\
        Weight decay & 0.1 \\
        Gradient clipping & 1.0 \\
        \midrule
        Gradient accumulation steps & 4 \\
        Effective prompt batch size & 4 \\
        Group size $G$ & 8 \\
        Sampled completions per update & 32 \\
        Sampling temperature & 1.0 \\
        KL coefficient $\beta$ & 0.04 \\
        Clipping parameter $\epsilon$ & 0.2 \\
        Maximum completion length & 256 tokens \\
        \bottomrule
    \end{tabular}
\end{table}

\begin{figure*}[t] 
    \centering
    \includegraphics[width=\linewidth]{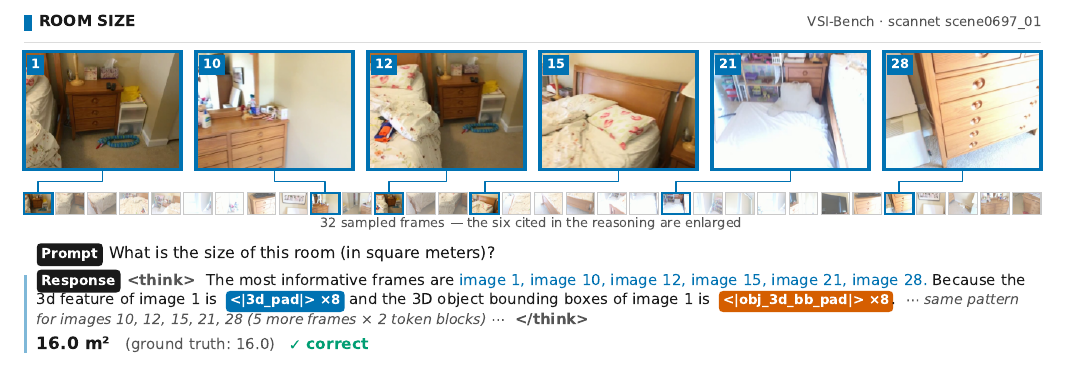} 
    \caption{Detailed prompt structure for stage 2 and 3 training. In this prompt style, we perform the reasoning based on six pre-selected frames and generate the answer. }

    \label{fig:prompts_appendix}
\end{figure*} 
\appsubsection{Prompting Structure}
\label{app:data_setup}

Different training stages require distinct prompt structures, as each stage emphasizes a different learning objective. As illustrated in Fig. 3, the complete set of spatial tasks is specified in the prompt at every stage. In Stage 1, the prompt contains the visual inputs together with an instruction to estimate the enabled spatial tasks. Structured task-token blocks are inserted into the assistant context, and the model is optimized exclusively using the auxiliary spatial-alignment losses, without language-model supervision.

In Stages 2 and 3, we introduce a structured, per-view chain-of-thought enclosed within \texttt{<think>...</think>} tags, which associates each selected image with its corresponding spatial-task tokens. By default, this reasoning sequence is provided through teacher forcing, and the model is trained to predict only the final answer. Alternatively, the reasoning trace can be included in the prediction target, allowing the model to generate it autoregressively. However, we find that explicitly generating the reasoning increases inference time without providing a meaningful performance improvement over teacher-forced reasoning.

The reasoning structure in Stages 2 and 3 uses only a subset of the input frames rather than all 32 frames. To identify the most informative views, we first trained a model to generate the reasoning trace and used CLIP similarities between the textual prompt and the 32 input frames to rank the frames by relevance. We additionally observed that, when answering questions from our primary benchmark, VSI-Bench, the trained model relied more frequently on certain frames than on others. Based on this observation, we selected the six highest-ranked frames (1, 10, 12, 15, 21, and 28) and used only these views for spatial reasoning during Stages 2 and 3. As illustrated in Fig.~\ref{fig:prompts_appendix}, the reasoning section of the prompt begins with the statement \texttt{The most informative frames are image 1, image 10, ...}, followed by reasoning over the corresponding 3D scene representations and 3D object bounding boxes.

\begin{table}[t]
\centering
\caption{Comparison between number of frames for reasoning in the prompt for the SenseNova-SI-1.3 + Space Tokens model.}
\label{tab:frames}

\begin{tabular}{lcc}
\toprule
\textbf{Number of Frames for Reasoning} & \textbf{VSI-Bench Avg.} \\
\midrule
0 (SenseNova-SI-1.3 baseline) & 67.6 \\
32 & 67.6 \\
4  & 68.4 \\
6  & 68.6 \\
\bottomrule
\end{tabular}%
\end{table}

We further analyze the effect of the number of reasoning frames using the model obtained after Stages 1 and 2. As shown in Tab.~\ref{tab:frames}, reasoning over all 32 frames achieves performance comparable to the original SenseNova-SI-1.3 checkpoint, which does not perform explicit spatial reasoning and therefore corresponds to zero reasoning frames. In contrast, using six selected frames yields the best performance among the evaluated configurations. These results suggest that reasoning over a compact set of informative views is more effective than processing the entire frame sequence indiscriminately. Many frames may depict uninformative or question-irrelevant regions, such as ceilings, the ground, or unrelated parts of the scene, thereby introducing noise into the reasoning process. Consequently, selecting a subset of relevant frames provides a more focused and effective basis for spatial reasoning, consistent with both intuition and our empirical findings.

\appsection{Additional Ablations}

\appsubsection{Token Replacement Test}
\label{app:trt}
To evaluate the utility of newly introduced tokens, \cite{Zhang_ablate_to_validate} proposed the Token Replacement Test (TRT), in which the input embeddings corresponding to the proposed tokens are perturbed before inference. The perturbation is done by replacing the token embedding with either a zero vector, a random vector, or a random vector sampled from the same distribution as the latent representation. If downstream performance deteriorates under this perturbation, the test suggests that the model relies on those token embeddings when producing its predictions.

\begin{table}[!tp]
    \caption{Token replacement test results for SenseNova-SI-1.3 + Space Tokens on VSI-Bench. Perturbing spatial token input embeddings negligibly impacts spatial understanding.}
    \label{tab:trt}
    \centering
    
    \begin{tabular}{lc}
        \hline
        \textbf{Mode} & \textbf{VSI-Bench Avg.} \\
        \hline
        unperturbed & 69.0 \\
        zero & 68.8 \\
        random & 68.6 \\
        random (same dist.) & 68.7 \\
        \hline
    \end{tabular}

\end{table}

    


\begin{table*}[!tp]
\centering
\caption{Attention masking results on VSI Bench. Removing attention from spatial tokens greatly impacts spatial understanding.}
\label{tab:attn_mask}

\resizebox{\textwidth}{!}{
\begin{tabular}{lcccccccccc}
\toprule
\multirow{2}{*}{Mode}
& \multirow{2}{*}{Avg.}
& \multicolumn{4}{c}{Numerical Questions}
& \multicolumn{4}{c}{Multiple-Choice Questions} \\
\cmidrule(lr){3-6}
\cmidrule(lr){7-10}
&
& Obj. Cnt.
& Abs. Dist.
& Obj. Size
& Room Size
& Rel. Dist.
& Rel. Dir.
& Route Plan
& Appr. Order \\
\midrule

No masking & 63.4 & 72.3 & 55.8 & 79.0 & 63.2 & 62.0 & 71.9 & 30.9 & 72.2 \\
Spatial token masking & 55.3 & 60.4 & 46.3 & 63.7 & 58.2 & 50.8 & 60.1 & 32.0 & 70.6 \\

\bottomrule
\end{tabular}
}

\end{table*}

\begin{table*}[t]
\centering
\caption{Accuracy (\%) on VSI Bench by checkpoint and input frames across min/max pixel settings. Best result per column in \textbf{bold}, second best \underline{underlined}.}
\label{tab:image-config}
\begin{tabular}{llcccc}
\toprule
& & \multicolumn{4}{c}{Min / Max pixels} \\
\cmidrule(lr){3-6}
Input frames & Checkpoint & 200{,}704 / 200{,}704 & 200{,}704 / 602{,}112 & 602{,}112 / 602{,}112 & 802{,}816 / 802{,}816 \\
\midrule
\multirow{3}{*}{32}
  & Base    & 67.6 & 67.6 & 68.0 & 68.0 \\
  & Stage 1\&2 & 67.9 & 68.7 & 68.8 & \underline{69.1} \\
  & GRPO    & 68.5 & \underline{69.0} & \textbf{69.3} & \textbf{69.3} \\
\midrule
\multirow{3}{*}{128}
  & Base    & 67.1 & 67.7 & 67.1 & 67.3 \\
  & Stage 1\&2 & \underline{68.6} & 68.8 & 68.6 & 67.7 \\
  & GRPO    & \textbf{68.9} & \textbf{69.2} & \underline{69.0} & 68.2 \\
\bottomrule
\end{tabular}
\end{table*}

While TRT is a useful diagnostic for methods trained on single-image inputs, such as CoVT~\cite{Qin_covt}, its interpretation becomes less clear in multi-image settings. In single-image models, each learned token has a fixed semantic role and its input embedding can directly indicate where the model should extract information. In contrast, multi-image models must additionally identify which image each token corresponds to. This association cannot be inferred from a constant input embedding alone, since the same learned token is reused across every image. Instead, image identity is specified through positional information elsewhere in the prompt.

For Space Tokens, this positional information is explicitly encoded in the reasoning template. For example, the phrase ``Because the 3D feature of image 5 is \textless|3d\_pad|\textgreater{} $\times$8'' informs the model that the subsequent spatial tokens correspond to the fifth input image. Consequently, the input embedding primarily identifies the token type, while the prompt structure determines which image-specific latent representation should be retrieved.

This distinction suggests that TRT primarily measures the model's sensitivity to perturbations of the \emph{input embedding}, rather than the usefulness of the latent representations computed from those tokens. Consistent with this interpretation, Mull-Tokens~\cite{Ray_mull}, which is also trained on multi-frame visual inputs, reports that TRT produces only negligible changes in performance on BLINK~\cite{fu2024blink}. We observe the same behavior for our Space Tokens SenseNova-SI-1.3 Stage 3 model when evaluated on VSI-Bench, as shown in Tab.~\ref{tab:trt}. The minimal performance degradation therefore should not be interpreted as evidence that the spatial token representations are unused; rather, it indicates that perturbing the input embeddings is an insufficient probe for evaluating latent representations in multi-image architectures.

To directly measure whether the information encoded in the spatial token representations contributes to model predictions, we instead propose a causal attention masking experiment. 

\appsubsection{Token Attention Masking}
\label{app:attn_mask}

Rather than perturbing the token embeddings before they are processed by the transformer, we instead intervene directly on the resulting latent representations. Specifically, we construct a causal attention mask that prevents the VLM from attending to the spatial tokens during answer generation while leaving all other computations unchanged.

The results of attention masking on the Space Tokens SenseNova-SI-1.3 Stage 3 model are reported in Tab.~\ref{tab:attn_mask}. Removing attention to the spatial tokens produces a substantial decrease in VSI-Bench performance, which is especially notable in the object size metric, which observes a 15.3 point drop. This demonstrates that the information encoded in these latent representations is causally required for accurate spatial reasoning. Unlike TRT, which probes sensitivity to the initial input embeddings, this experiment directly evaluates whether the learned representations themselves are utilized by the VLM.

Since Flash Attention 2 does not support arbitrary attention masks, these experiments are performed using eager attention. Flash Attention 2 and eager attention implement the same scaled dot-product attention operation, but differences in floating-point accumulation can affect token probabilities and subsequent autoregressive predictions. This produces a reduction in absolute performance compared with Flash Attention 2, which is why the no masking experiment sees a drop from $69.0$ to $63.4$. However, this effect is common to both the masked and unmasked evaluations and therefore does not affect the conclusions of the study.

\appsubsection{Evaluation Image Configurations}
\label{app:eval_image}

Providing additional visual information at test time has been shown to improve VLM performance on spatial benchmarks \cite{Cai_EASI}. We conduct ablations along two axes: the number of input frames and the minimum/maximum pixels per image. Increasing the number of input frames increases the effective sampling rate and provides more temporal context. Increasing the minimum/maximum pixels per image maintains higher image resolution, which preserves finer image features. It is worth noting that the frames selected for reasoning were adjusted accordingly to account for the increased number of input frames. The results of the sweep are reported in Tab. ~\ref{tab:image-config}. Independently increasing along each axis yields an improvement in model performance, except for larger pixel counts on 128 frames. This degradation may be due to the interaction between frames and pixel count, which increases the number of visual tokens. The model was trained on smaller context lengths which may result in worse performance in cases where excess visual tokens dilute the context window of relevant information.
\FloatBarrier

\end{document}